\documentclass[conference]{IEEEtran}
\usepackage{authblk}
\usepackage{mathptmx}
\usepackage{fancyhdr}
\usepackage{amsmath}
\usepackage[normalem]{ulem}
\usepackage[hyphens]{url}
\usepackage[sort,nocompress]{cite}
\usepackage[final]{microtype}
\usepackage[keeplastbox]{flushend}
\usepackage[linesnumbered, ruled, noend]{algorithm2e}
\usepackage{enumitem}
\usepackage{array,multirow}
\newcolumntype{?}{!{\vrule width 1pt}}
\usepackage{booktabs}
\usepackage{tcolorbox}
\usepackage{comment}
\usepackage{xcolor}
\usepackage{caption}
\usepackage{atbegshi,picture}
\usepackage{eso-pic}
\usepackage{lipsum}
\usepackage[export]{adjustbox}
\newtheorem{theorem}{Theorem}

\PassOptionsToPackage{pdfpagelabels=false}{hyperref} 

\usepackage[hyphens]{url}

\def\BibTeX{{\rm B\kern-.05em{\sc i\kern-.025em b}\kern-.08em
    T\kern-.1667em\lower.7ex\hbox{E}\kern-.125emX}}
\title{FAST: DNN Training Under Variable Precision Block Floating Point with Stochastic Rounding}
\author[1]{Sai Qian Zhang}
\author[2]{Bradley McDanel}
\author[1]{H.T. Kung}
\affil[1]{Harvard University}
\affil[2]{Franklin and Marshall College}

\begin{document}

\AddToShipoutPictureBG*{%
  \AtPageUpperLeft{%
    \hspace{0.9\paperwidth}%
    \raisebox{-2.5\baselineskip}{%
      \makebox[0pt][r]{\begin{tabular}{l}Preliminary version. The final version is to appear\\in the 28th IEEE Int Symp on
High-Performance \\ Computer Architecture (HPCA-28), February 2022\end{tabular}}
}}}%

\maketitle
\thispagestyle{plain}
\pagestyle{plain}

\begin{abstract}
Block Floating Point (BFP) can efficiently support quantization for Deep Neural Network (DNN) training by providing a wide dynamic range via a shared exponent across a group of values. In this paper, we propose a Fast First, Accurate Second Training (FAST) system for DNNs, where the weights, activations, and gradients are represented in BFP. FAST supports matrix multiplication with variable precision BFP input operands, enabling incremental increases in DNN precision throughout training. By increasing the BFP precision across both training iterations and DNN layers, FAST can greatly shorten the training time while reducing overall hardware resource usage. Our FAST Multipler-Accumulator (fMAC) supports dot product computations under multiple BFP precisions. We validate our FAST system on multiple DNNs with different datasets, demonstrating a 2-6$\times$ speedup in training on a single-chip platform over prior work based on \textbf{mixed-precision or block} floating point number systems while achieving similar performance in validation accuracy.
\end{abstract}

\section{Introduction}
Custom floating point (FP) formats, such as Google's bfloat16~\cite{bfloat} and Nvidia's TensorFloat 32~\cite{tf32}, are increasingly replacing IEEE 754 32-bit floating point (FP32) for DNN training. These formats more efficiently fit the empirical distribution of DNN weight, data, and gradient values, leading to a smaller hardware footprint for the multiplier-accumulator (MAC) unit. However, these formats are still significantly more expensive to implement than fixed point (INT) formats of similar bitwidths due to mantissa alignments which are required for each floating point MAC operation.

By comparison, Block Floating Point (BFP)~\cite{Wilkinson_64} formats offer a middle ground between FP and INT formats, by enforcing that a group of values share a common exponent while maintaining individual mantissas. This constraint enables BFP to achieve higher efficiency than FP for dot product (DP) computations for three reasons. First, mantissa alignments are only required on input values in each BFP group as opposed to after each FP multiplication. Second, there is only one exponent addition between each group as opposed to each FP multiplication. Therefore, performing DP computations in BFP can lead to a significant improvement in training efficiency.

In this work, we propose a \textbf{F}ast First, \textbf{A}ccurate \textbf{S}econd \textbf{T}raining (\textbf{FAST}) system for variable precision BFP DNN training. Here, variable precision means that (1) the system efficiently supports BFP formats across a range of mantissa widths during training and (2) dot products between BFPs with different mantissa widths is permitted. To support our system, we have designed a FAST multiplier-accumulator (fMAC) which operates on n-bit chunks of mantissas across two groups of BFP numbers being multiplied. Throughout this paper, we use 2-bit chunks. Sub-dividing the computation into 2-bit chunks allows the same fMAC to implement arithmetic operations involving higher precision mantissas by simply running multiple passes of the fMAC. For instance, multiplying two groups with 2-bit and 4-bit BFP mantissas translates to $\frac{2}{2}\times \frac{4}{2} = 2$ passes. The rate at which our FAST system performs dot product computations is based on the BFP precision of the two vectors being multiplied. 

\begin{figure}
\centering
\includegraphics[width=\columnwidth]{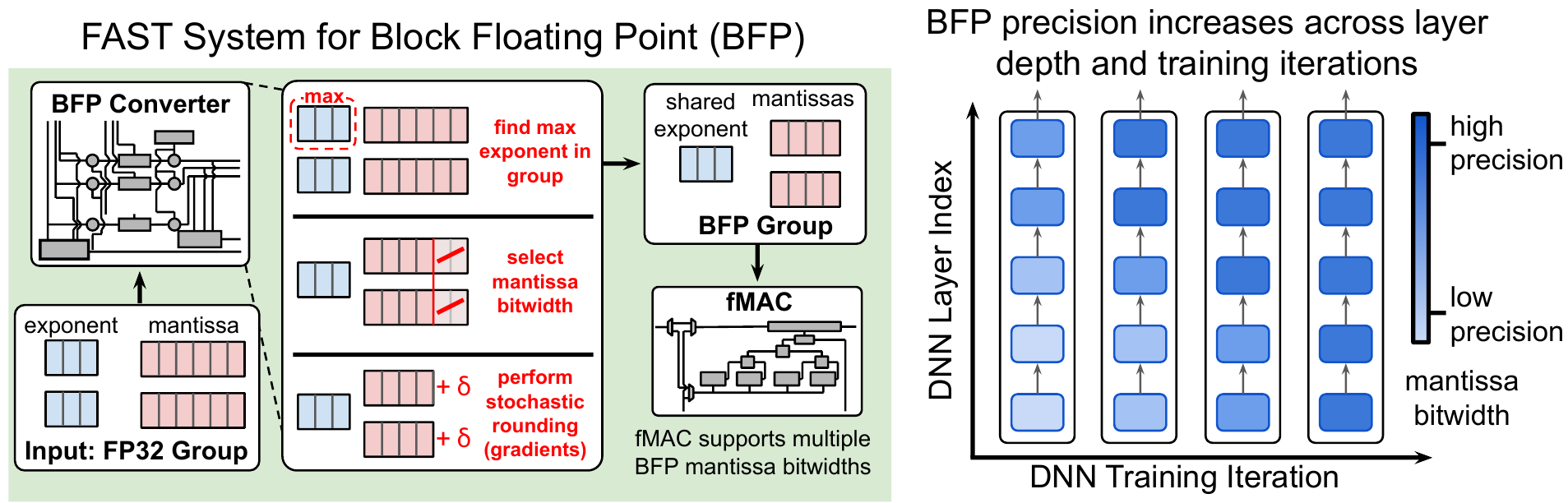}
\caption{(left) Overview of the proposed FAST system for DNN training with variable precision BFP with stochastic rounding. A group of FP32 values are converted to a BFP group with a selected mantissa width based on the current error tolerance (see Algorithm~\ref{alg:adaptive-fast-training}). (right) FAST incrementally increases the BFP precision across both layer depths and training iterations. See Figure~\ref{fig:heatmap} on corresponding empirical data obtained from FAST training of Resnet-18 on ImageNet.}
\label{fig:overview}
\end{figure}

With FAST, we propose a DNN training regime that starts with low-precision BFP and increases the precision of weights, data, and gradients over the course of training. Figure~\ref{fig:overview} presents an overview of how FAST can accelerate training via low-precision operations. The left side of the figure provides a sketch of the FAST system, which supports FP32 to BFP conversion for a range of mantissa bitwidths and stochastic rounding for gradients (to maintain training stability under low-precision BFP). The FAST compute engine consists of a systolic array~\cite{kung1982systolic} of fMAC units, which efficiently supports BFP dot products with varying mantissa bitwidths. The right side of the figure shows how the precision of the mantissa field in BFP for a DNN increases across DNN layers and over training iterations. While each DNN layer in the figure is presented with a single precision, in practice the precision of the weight, data, and gradient tensors in each layer are selected independently for a given iteration (see Figure~\ref{fig:heatmap} for how this precision selection works). Using this approach, FAST is able to achieve this result on (1) CNNs for ImageNet~\cite{deng2009imagenet}, (2) Transformers for the IWSLT14 German-English benchmark~\cite{iwslt-benchmark}, and (3) YOLOv2~\cite{redmon2017yolo9000} for the PASCAL VOC2012~\cite{everingham2011pascal} dataset.

In FAST, we use (1) BFP for variable precision training and (2) BFP with stochastic rounding. BFP is an old idea dating back as early as the 1960s (see, e.g.,~\cite{Wilkinson_64}), and there has been recent literature demonstrating the advantages of using BFP in DNN inference~\cite{rouhani20bfp} and training~\cite{drumondbfp18}. We believe that our ideas of (1) and (2) are novel. For (1), we point out the convenience in Section~\ref{sec:hw-eval} of implementing variable precision hardware for BFP. For (2), we note in Section~\ref{sec:bfp:dnn-dist} that using stochastic rounding in conjunction with BFP is critical to model accuracy, especially when using BFP for gradients with low-precision mantissa (e.g., 2 or 4 bits). In Section~\ref{sec:bfp:sr} we provide an analysis of the reasons for using stochastic rounding in BFP.
The main contributions of the paper are:
\begin{itemize}
    \item The \textit{FAST variable precision training algorithm} for efficient DNN training. The proposed solution reduces total training time by adaptively selecting the optimal precision for weights, data, and gradients in every DNN layers at each iteration.
    \item Our proposed use of (1) BFP for variable precision training and (2) BFP with stochastic rounding. We provide a novel analysis of the impact of applying stochastic rounding to weight gradients on the loss in gradient decent (Theorem~\ref{thm:rounding} in Section~\ref{sec:bfp:sr}).
    \item A modular architecture consisting of \textit{FAST multiplier-accumulator} (fMAC) for groups of BFP values. fMAC operates on chunks of BFP mantissas (e.g., 2-bit chunks) to support variable-width mantissas in 2-bit increments.
\end{itemize}
Table~\ref{tab:terms} lists the terminology and notation used in the paper.
\begin{table}[]
\centering
\caption{Terminology and notation used in the paper.}
\begin{adjustbox}{width=0.9\columnwidth,center}
\begin{tabular}{|c|l|}
\hline
Notations   &  \multicolumn{1}{c|}{Explanation}  \\ \hline
\textbf{BFP} & BFP-quantized values  \\ \hline
\textbf{DP} & Dot product  \\ \hline
\textbf{FP} & Floating point. Assume FP32, unless otherwise stated  \\ \hline
\textbf{INT} & Fixed point integer  \\ \hline
\textbf{UQ} & Uniform quantization  \\ \hline
\textbf{SR} & Stochastic rounding   \\ \hline
$g$ & Group size for BFP. Assume g=16, unless otherwise stated.  \\ \hline
$e$ & Exponent bitwidth for FP and BFP  \\ \hline
$m$ & Mantissa bitwidth for FP, BFP, and INT  \\ \hline

\end{tabular}
\end{adjustbox}
\label{tab:terms}
\end{table}
\vspace{0.5em}


\section{Background and Related Work}
\label{sec:bg}

In Section~\ref{sec:bg:fp-formats}, we provide an overview of number formats for DNN training and inference. Section~\ref{sec:bg:dnn-training} provides an overview of the computation dataflow of DNN training. Finally, in Section~\ref{sec:bg:dnn-training-archs}, we review related work on hardware accelerators for DNN training.

\subsection{Number Formats for DNNs}
\label{sec:bg:fp-formats}
\begin{figure}[t]
\centering
\includegraphics[width=0.95\columnwidth]{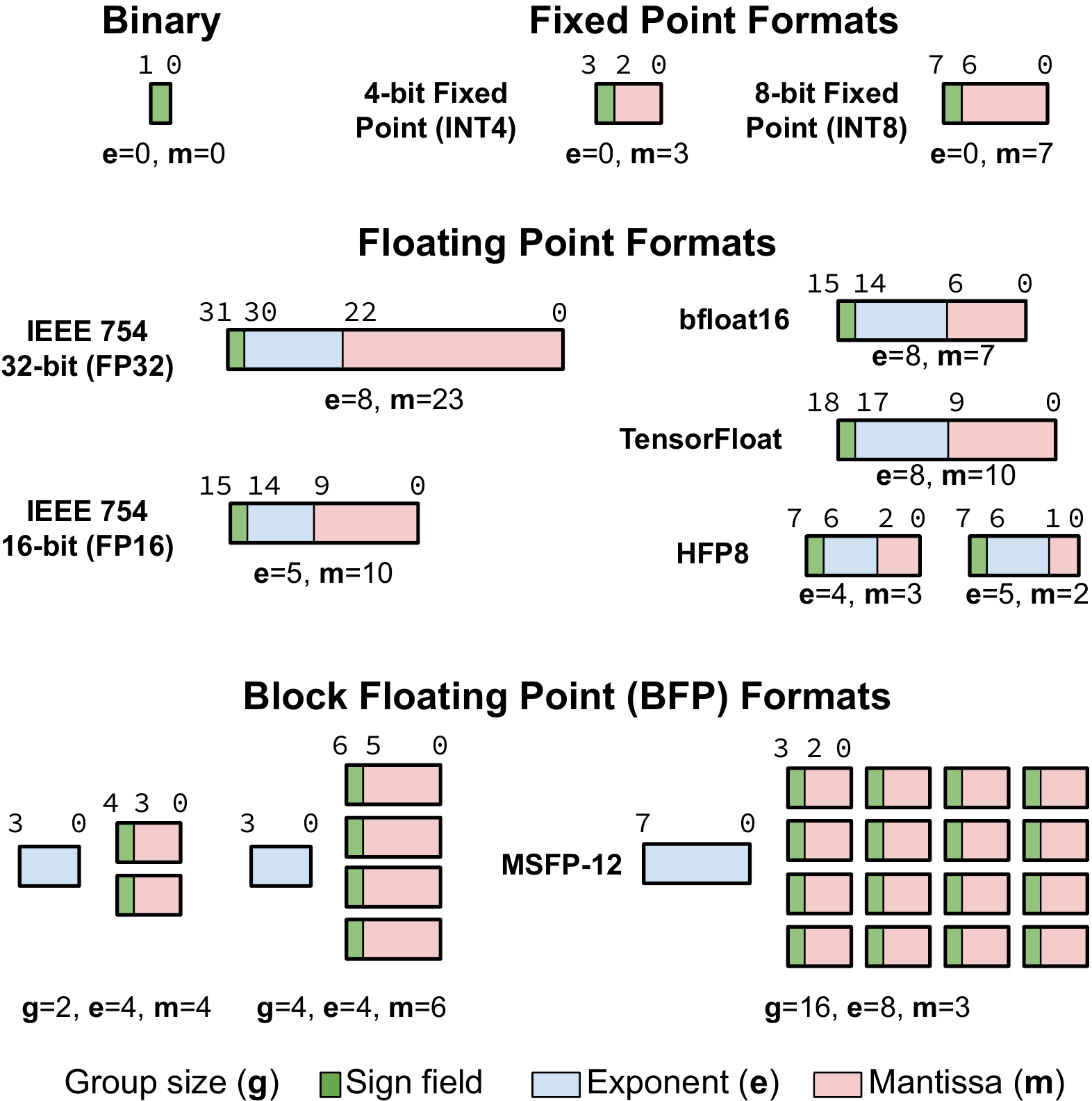}
\caption{Number formats commonly used for DNN training and inference. Fixed point formats (top) are often used for inference. Floating point formats (middle) are used for training, with bfloat16 and TensorFloat replacing IEEE 754 in most cases by increasing the number of exponent bits to widen the dynamic range. BFP (bottom) fits between fixed and floating point by sharing a single exponent across the group values.}
\label{fig:fp-formats}
\end{figure}

As the majority of computation in both DNN training and inference are dot products, the formats for their underlying numbers has been extensively studied to make this computation efficient. Figure~\ref{fig:fp-formats} divides various formats into three groups: fixed point (top), floating point (middle), and BFP (bottom). The number of exponent bits (e) and mantissa bits (m) are provided for each format.

Fixed point formats do not have an exponent field, which reduces the dynamic range that can be represented but simplifies the hardware. The use of fixed point formats for DNN training and inference has been well explored~\cite{courbariaux2014training,courbariaux2015binaryconnect,gupta2015deep,zhu2016trained,hubara2017quantized,park2017weighted,kapur2017low,banner2018scalable,jacob2018quantization,bilaniuk2019bit,kung2019packing}. The smallest format is a 1-bit binary representation (upper left of Figure~\ref{fig:fp-formats}) used by binarized neural networks which has no exponent or mantissa bits. Floating point formats have a larger dynamic range than fixed point, making them more amenable to wider dynamic range of gradients in DNN training~\cite{chmiel2020neural}. IEEE 754 32-bit IEEE floating point or FP32 (middle left of Figure~\ref{fig:fp-formats}) is the conventional format for DNN training. Mixed precision training operates on some tensors in a higher precision and other tensors in a lower precision. For instance, Nvidia Mixed Precision (MP)~\cite{micikevicius2018mixed} proposed to perform most computations in the forward pass in FP16, while keeping an FP32 copy of the weights for updating during training. Custom floating point formats like bfloat16~\cite{bfloat}, TensorFloat~\cite{tf32} and HFP8~\cite{sun2019hybrid} (middle right of Figure~\ref{fig:fp-formats}) operate in a similar mixed precision regime and have been shown to work as well as FP32 for training accurate DNNs. For example, HFP8 performs forward pass computations using 8-bit FP with one bit sign, 4-bit exponent, 3-bit mantissa (1-4-3) and backward pass computations with one bit sign, 5-bit exponent and 2-bit mantissa (1-5-2).




While BFP represents a promising direction for improving the efficiency of DNN training as middle ground between fixed and floating point, there has been a small amount of prior work in designing efficient hardware to support it. Flexpoint~\cite{koster2017flexpoint} proposed a BFP format with a 16-bit mantissa (m=16) and a 5-bit shared exponent across an entire tensor. Our paper focuses on how to adjust the BFP mantissa bitwidth adaptively (e.g., m = 2 or 4) during training to reduce training time and power consumption. Drumond et al.~\cite{drumondbfp18} proposed to use a large BFP group size of 576 (a 2D tile of size $24\times24$) which requires a wide mantissa bitwidth of m=12 to achieve good accuracy. Compared to these prior approaches, we show in Section~\ref{sec:sw-eval:float-comp}, that training under INT with similar number of bits (e.g., 12 bits) also has good performance, which suggests that BFP has little advantages over INT12 for such large tiles.  Additionally, they did not provide a detailed hardware design for the implementation of BFP computation.

More recently, Microsoft proposed a BFP format (MSFP-12 in lower right of Figure~\ref{fig:fp-formats}) for DNN inference via post-training quantization~\cite{rouhani20bfp} on their Project Brainwave FPGA cloud platform\cite{fowers2018brainwave}. In our paper, by using BFP-aware DNN training instead of post-training quantization, we are able to use a smaller exponent width of 4 instead of 8 while achieving similar inference accuracy. For training, as we show in Figure~\ref{fig:asic-results}, via variable precision BFP, FAST reduces the training time and power consumption compared to the previously mentioned training based on floating point or BFP formats.

\subsection{Matrix Computation of DNN Training}
\label{sec:bg:dnn-training}

\begin{figure}
\centering
\includegraphics[width=\columnwidth]{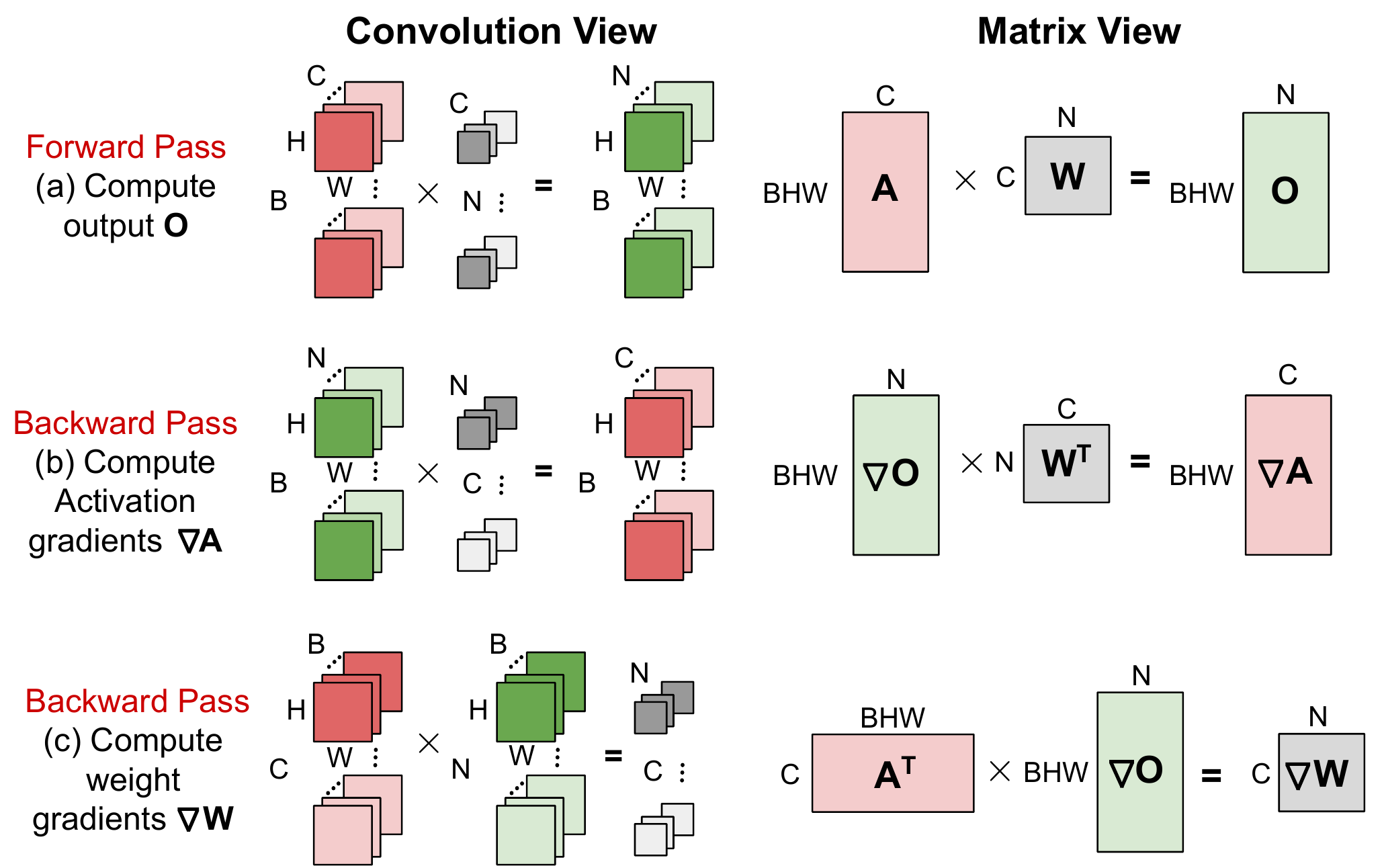}
\caption{The forward and backward pass steps for a single layer of DNN training represented in a convolution view (middle) and matrix operation view (right) with matrix dimensionalities shown. The kernel size of the convolution filters is 1x1 for presentation simplicity.}
\label{fig:dnn-training-computation}
\end{figure}

Each iteration of DNN training on a mini-batch consists of a forward pass to compute a loss and a backward pass to update the DNN weights with gradients computed from the loss. Figure~\ref{fig:dnn-training-computation} illustrates all of the matrix computations required for both forward and backward passes for one convolutional layer in a CNN (fully connect layers operate in a similar manner). Both the convolutional view and the corresponding matrix operation view are presented. During the forward pass, the input activations ($A$) are convolved with the layer weights ($W$) to compute the output ($O$) as depicted in Figure~\ref{fig:dnn-training-computation}a. Then, $O$ will be passed through normalization and a non-linear activation function, to become the input activations for the next layer. 

During the backward pass, two convolutions are performed at each layer. Figure~\ref{fig:dnn-training-computation}b shows how the output gradients $\nabla O$ are convolved with the transposed weights $W^{T}$ to compute the activation gradients $\nabla A$, which will be passed to the layer below. In Figure~\ref{fig:dnn-training-computation}c, the transposed input activations $A^{T}$ are convolved with the output gradients $\nabla O$ in order to compute the weight gradients $\nabla W$. These weight gradients $\nabla W$ are then added in an elementwise fashion with the weights W in order to compute the updated weights $W'$. 

\subsection{Accelerators for DNN Training}
\label{sec:bg:dnn-training-archs}
Previous work on accelerating the DNN training has focused on leveraging the sparsity present in weights and activations~\cite{mahmoud2020tensordash,zhang2019eager,yang2020procrustes,choi2020energy}. TensorDash~\cite{mahmoud2020tensordash} accelerates the DNN training process while achieving higher energy efficiency via eliminating the ineffectual operations resulted from the sparse input data. Eager Pruning~\cite{zhang2019eager} and Procrustes~\cite{yang2020procrustes} improve DNN training efficiency by co-designing the training algorithm with the target hardware platform (``hardware-aware training''). Insignificant DNN weights are pruned in the middle of the DNN training process and ineffectual operation involving zero weights can be eliminated without impacting the final accuracy. In comparison, our approach applies BFP to dynamically adjust the precision of DNN training, which is in orthogonal to these methods which exploit value-level sparsity.

Multi-precision methods of reducing the computation of DNNs have been explored in the literature. Stripes~\cite{judd2016stripes} multiplies two 16-bit integers by only adding those shifted multiplicands corresponding to the nonzero bits during DNN inference. In the work, we use short 2-bit chunks of mantissas, making a straightforward single clock bit-parallel implementation efficient compared to a bit-serial approach for DNN training. Lee et al.~\cite{lee20197} proposed using fine-grained mixed precision (FGMP) of FP8-FP16, which represents some parts of a tensor in FP8 and other parts in FP16. FAST uses both 2-bit and 4-bit mantissas, and iterates like FGMP on the low-bitwidth hardware for performing the high-bitwidth arithmetics. FAST performs an integer MAC for each partial product in a BFP group, rather than an FP MAC as in FGMP.

\section{Overview of BFP with Stochastic Rounding}
\label{sec:bfp}

In this section, we provide an overview of how we use BFP under stochastic rounding (SR) in FAST to facilitate efficient and accurate DNN training.

\subsection{Quantization from FP to BFP under Stochastic Rounding}
\label{sec:bfp:conversion}
\begin{figure}
\centering
\includegraphics[width=\columnwidth]{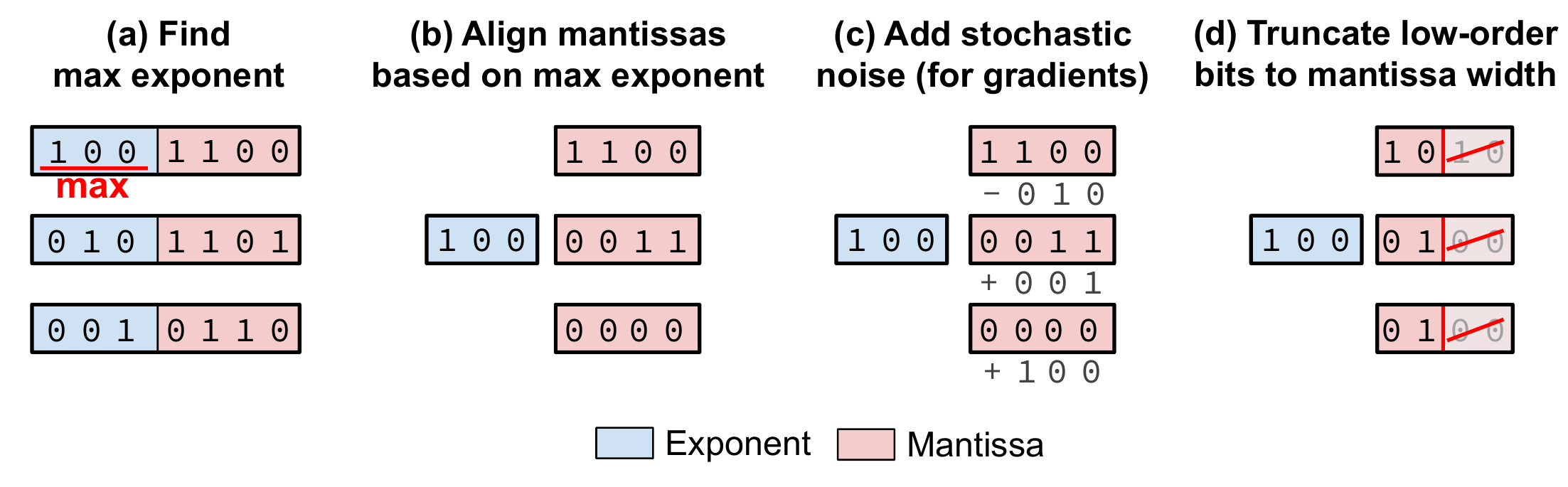}
\caption{(a) The max exponent across all values in the group is found. (b) The mantissas fields are aligned based on differences to the max exponent. (c) Stochastic noise is added to each mantissa (for gradients only). (d) The low-order mantissa bits are truncated to a fixed width.} 
\label{fig:fp-to-bfp}
\end{figure}

Figure~\ref{fig:fp-to-bfp} shows this quantization process for converting a group of three FP values. First, in Figure~\ref{fig:fp-to-bfp}a the largest exponent in the group is found, which becomes the shared exponent for the group. Then, in Figure~\ref{fig:fp-to-bfp}b, the mantissas of each value are aligned based on the difference between the exponent of each value and the max exponent. Next, in Figure~\ref{fig:fp-to-bfp}c, stochastic noise is added for gradients (critically important for low bitwidth mantissas). Finally, in Figure~\ref{fig:fp-to-bfp}d, the low-order mantissa bits are truncated to a specified mantissa bitwidth. 

\subsection{Dot Product under FP, INT, and BFP}
\label{sec:bfp:dot-product}
In this section, we discuss the conversion and computation costs of dot product (DP) under three number formats (BFP, INT, and FP). We argue that BFP DP is less costly than FP DP due to its use of shared exponents, and BFP DP is less costly than INT DP because the former can achieve the same accuracy as the latter with a smaller mantissa bitwidth. 

Consider the DP of two vectors of length $g$ (e.g., an activation vector and a weight vector). The DP computation can be broken into two parts: (Part M) $g$ integer multiplications for computing $g$ partial products and (Part A) accumulation of $g$ partial products resulting from part M.

\subsubsection{FP Conversion Cost (BFP DP versus INT DP)}
\label{sec:bfp-dp-cost}
Before performing BFP DP, we must first convert values in the two FP input vectors into BFP values. 
After the DP is computed, we need to convert the result to FP to add to an accumulation across BFP groups. This conversion operation is an FP normalization that involves bit shifts of the mantissa. 

For the INT DP, we need to perform similar conversions between FP and INT. Suppose that We use conventional uniform quantization (UQ) for the conversion from FP to INT. Since the scale factor is in FP, the INT conversion cost is much higher than the BFP conversion. The conversion of the DP result from INT to FP is an FP normalization, like the conversion of the BFP DP result to FP. 
\begin{figure}
\centering
\includegraphics[width=\columnwidth]{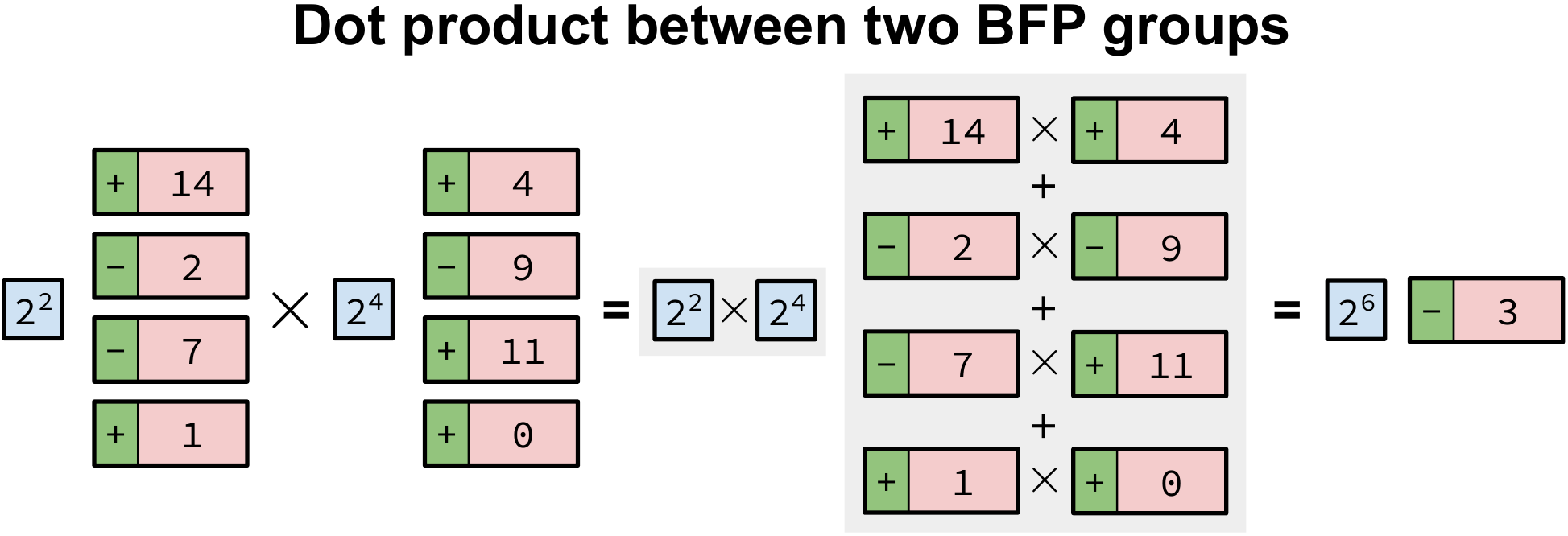}
\caption{The dot product between two BFP groups can be decomposed into fixed point multiplication between each pair of values in the groups and exponent addition between the two shared exponents.} 
\label{fig:bfp-dot-product}
\end{figure}
\subsubsection{Computation Cost (BFP DP versus FP DP)}
\label{sec:computation-cost-compare}
Figure~\ref{fig:bfp-dot-product} illustrates the dot product between two BFP groups of size $g$ = 4. We see that BFP DP costs substantially less than FP DP for three reasons. First, for part M, BFP DP just needs to perform one exponent addition on the shared exponents of the two input vectors. In contrast, FP DP requires $g$ exponent additions. Additionally, unlike FP DP, BFP DP does not perform FP normalization after each of the $g$ multiplications. Finally, for part A, unlike FP DP, BFP DP does not need to align partial products, as they are already aligned.

\subsubsection{Computation Cost (BFP DP versus INT DP)}
BFP can have a much smaller mantissa bitwidth $m$ (e.g., $m$ = 4) than INT (e.g., $m$ = 12) while achieving similar classification accuracy (Table~\ref{tab:arith-acc}).  Since the computational complexity of fixed point multipliers scales in a quadratic fashion with bitwidth, for part M, BFP DP costs much less than INT due to the reduced $m$. But, BFP does incur a relatively small cost compared to INT for adding the shared exponents between the BFP groups.

\subsection{BFP Exponent and Mantissa Bitwidths}
\label{sec:bfp:dnn-dist}
The number of exponent and mantissa bits in BFP play different roles in determining the amount of quantization error after conversion from FP to BFP. If the shared exponent bitwidth is too small, then it may not be able to represent numbers in the dynamic range of the group. If the mantissa bitwidth is too small, then some values with smaller exponents in a BFP group will have all mantissa bits shifted out of range resulting in data loss (see the third value in Figures~\ref{fig:fp-to-bfp}a and \ref{fig:fp-to-bfp}b with $m$ = 2). 



\begin{figure}
\centering
\includegraphics[width=0.97\columnwidth]{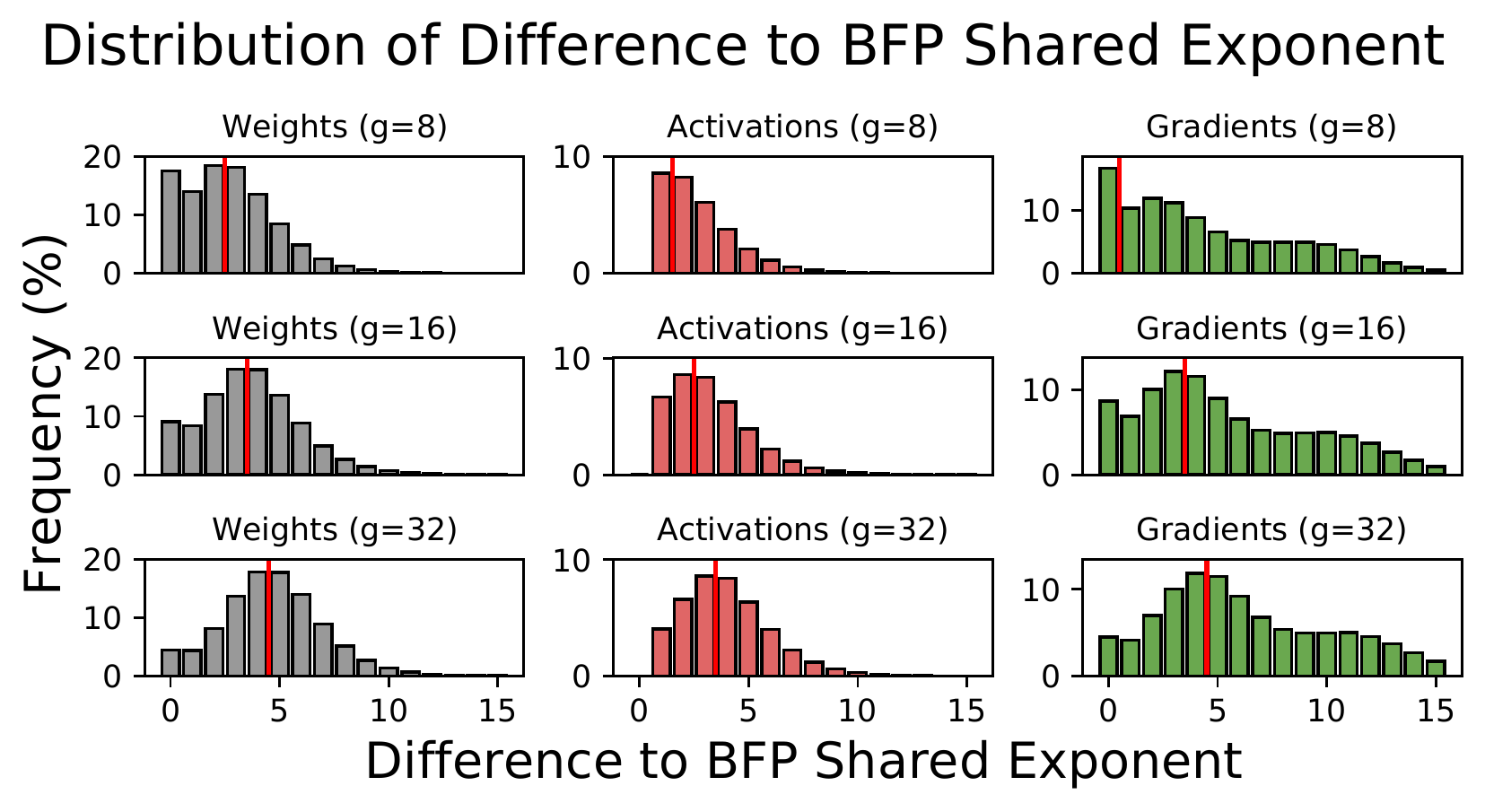}
\caption{The distribution of the difference between a BFP shared exponent (for three group sizes $g$ = 8, 16 and 32) and exponents of all other values in the group before mantissa alignment for ResNet-18 at layer 10. A larger difference leads to an increased quantization error (truncation error); see Figure~\ref{fig:fp-to-bfp}b.}
\label{fig:bfp-group-diff}
\end{figure}

Figure~\ref{fig:bfp-group-diff} presents the distribution of the difference in exponents between the maximum exponent in a group and all other exponents for three different group sizes (g = 8,16,32). The weight, data, and gradient tensors are taken from layer 10 in ResNet-18 at the halfway point of the training on ImageNet (other layers and DNNs generally follow the same trend). The difference dictates the amount of shifting required to align mantissas as depicted in Figures~\ref{fig:fp-to-bfp}b. Large differences lead to large worst-case quantization errors. When the difference is larger than the mantissa bitwidth, all bits will be truncated. Compared to the weights and activations, the gradients have a much wider exponent disparity, leading to a larger quantization error. This is why stochastic rounding (SR) for gradient computations, as depicted in Figure~\ref{fig:fp-to-bfp}c, is essential to achieve high accuracy when using a low number of mantissa bits. We notice that the mass of each distribution moves to the right as the group size $g$ increases, as indicated in the positions of the red vertical line. Thus, increasing the $g$ value will increase truncation errors for the same mantissa bitwidth. In this paper, we set $g$ at 16 unless otherwise stated.

\subsection{Stochastic Rounding (SR) of Gradients in Gradient Descent}
\label{sec:bfp:sr}
In this section, we present an analysis and illustrations on the working of SR in gradient descent for DNN training. The analysis shows that for low-precision BFP with small mantissa bitwidth \textit{m}, applying stochastic rounding to gradients, as illustrated in Figure~\ref{fig:fp-to-bfp}c, can minimize the impact of rounding on gradient descent performance.

\begin{figure}
\centering
\includegraphics[width=\columnwidth]{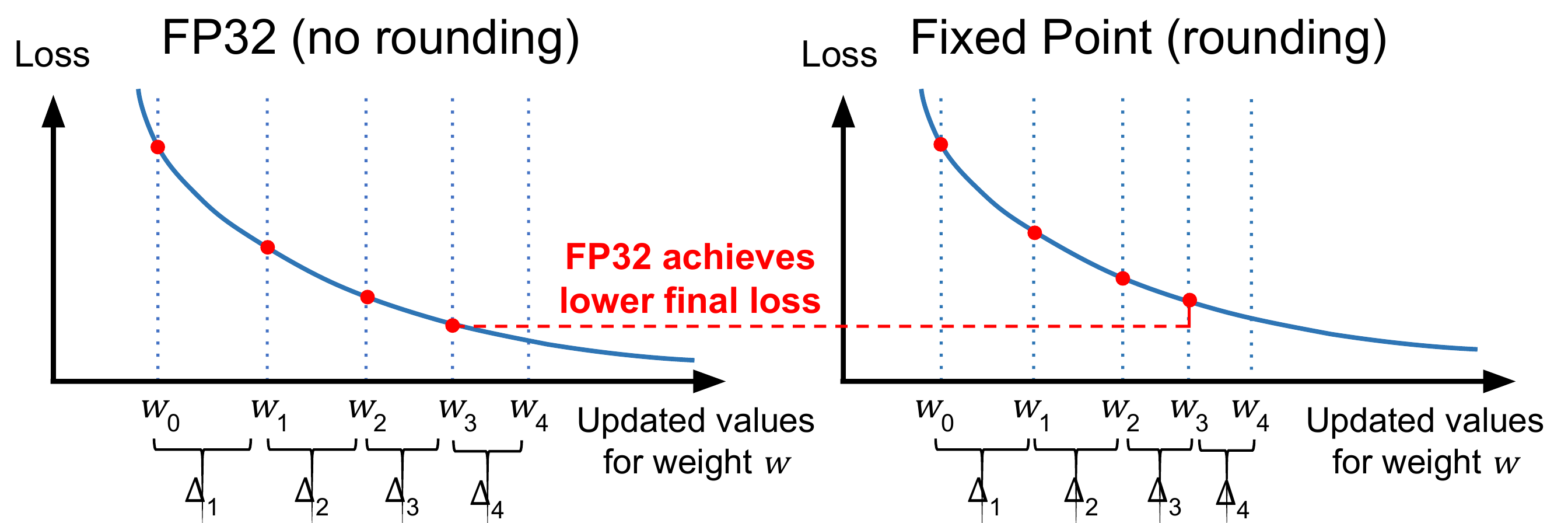}
\caption{(left) Weight $w$ is updated over 4 iterations using gradients $\Delta_1, \Delta_2, \Delta_3, \Delta_4$ computed in FP32 leading to a decrease in loss. (right) Suppose that rounding of weight gradients biases rounding each gradient down to smaller values and thus smaller $\Delta$ values. Then, these roundings lead to higher loss over multiple iterations.}
\label{fig:sr-explanation-a}
\end{figure}

Let $E$ be the training loss of a DNN (e.g., $E$ can be based on cross entropy). Without loss of generality we assume in this analysis that the learning rate $\eta$ is 1. We consider use of stochastic gradient descent (SGD) to minimize $E$ using multiple rounds of iterations. Consider any specific parameter $w$ of the neural network. For each iteration $i$, the backpropagation algorithm computes the partial derivative $\nabla_i = -\frac{\partial E}{\partial w_{i}}$ and updates w with the following rule:
\begin{align*}
    w_{i+1} = w_i + \eta\nabla_i \enspace\enspace\enspace   
    \text{or} \enspace \enspace \enspace \Delta_i = \eta\nabla_i
\end{align*}
\noindent where $\eta$ is the learning rate and $\Delta_i = w_{i+1} - w_i$. We refer to a collection of multiple iterations over all training data as an epoch. In Figure~\ref{fig:sr-explanation-a} (left), we illustrate four iterations of updating $w$ from its initial value $w_0$ to $w_1$, $w_2$, $w_3$ and $w_4$, using full-precision floating point numbers FP32 without rounding.

\subsubsection{Impact of rounding on weight updates}
In Figure~\ref{fig:sr-explanation-a} (right), we consider the scenario when we perform training iterations using fixed point integers quantized from FP32. The diagram illustrates that if the gradient at each of the four iterations is rounded down, then the total weight increments $\Delta_1 + \Delta_2 + \Delta_3 + \Delta_4$ will be reduced leading to a higher loss, compared to the FP32 case without rounding. This is because the gradient $\nabla$ at each red dot corresponding to an iteration is rounded down to a smaller value, causing a smaller weight increment $\Delta$.

\subsubsection{Use of stochastic rounding to minimize impact of rounding on weight updates}
We use SR to minimize the impact of rounding on weight updates and the corresponding reduction on loss.

\begin{theorem}
If the gradient $\nabla$ remains the same over iterations, then SR is expected to yield the same total weight increments as FP32 without rounding, assuming that the stochastic noise used by SR is full precision.
\label{thm:rounding}
\end{theorem}


The assumption that gradients stay the same is just to simplify the explanation below. The same argument can derive the \textit{expected} increment on a weight \textit{w} over an iteration based on the \textit{expected} gradient value for that iteration, without having to make the assumption of constant gradients.

\begin{figure}
\centering
\includegraphics[width=\columnwidth]{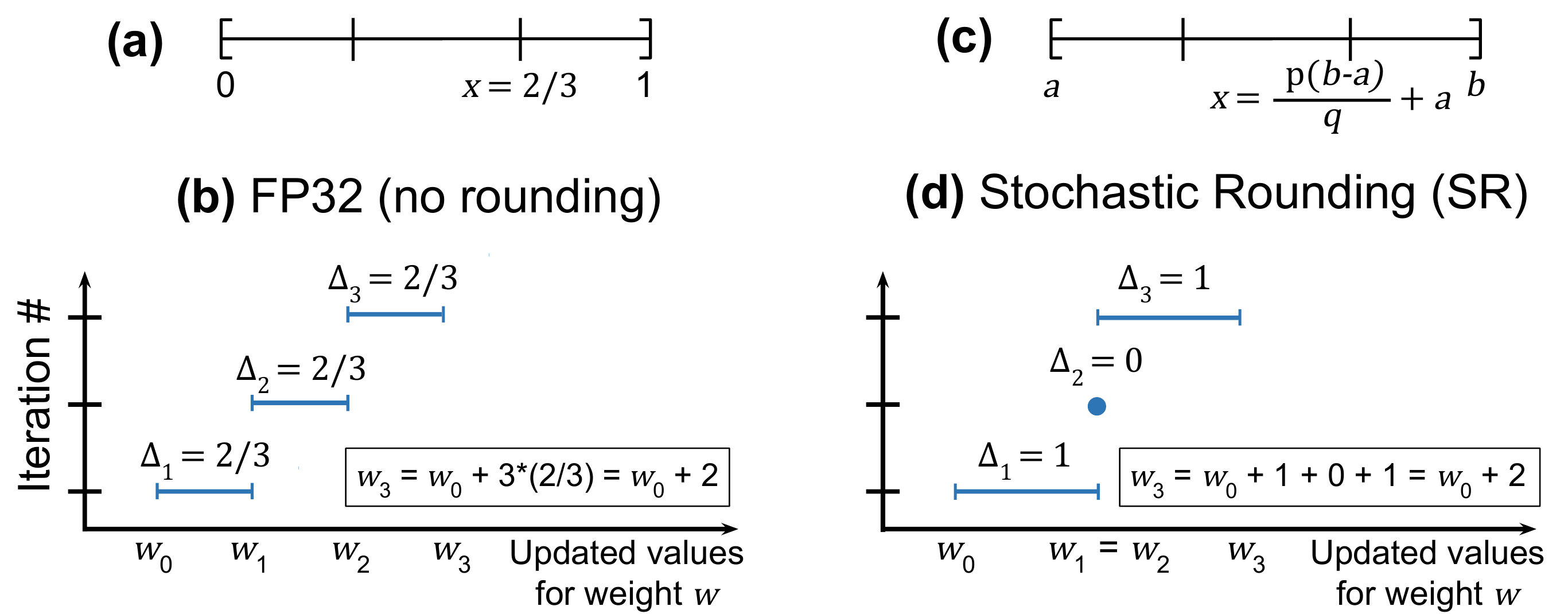}
\caption{(a) Gradient $x = 2/3$ in quantization decision interval $[0, 1]$. (b) In FP32, each iteration uses FP32 gradient $\Delta$ to update $w$. (c) General decision interval $[a, b]$ under SR. (d) SR is expected to increment $w$ the same amount as in (b) by rounding up to 1 twice and down to 0 once.}
\label{fig:sr-explanation-b}
\end{figure}

To explain Theorem~\ref{thm:rounding}, we first consider a simple case where we quantize a gradient $x$ ($= 2/3$) in a quantization decision interval $[0, 1]$, as depicted in Figure~\ref{fig:sr-explanation-b}a. Under SR, $x$ is rounded to 0 and 1 with probability 1/3 and 2/3, respectively, reflecting the distance of $x$ to each endpoint. Note that over 3 iterations, $x$ is expected to round down to 0 once and round up to 1 twice. Figure~\ref{fig:sr-explanation-b}d illustrates that $x$ is rounded to $1$, $0$ and $1$ for iteration 1, 2 and 3, respectively. We note that SR increments $w_0$ by the same amount (i.e., in Figure~\ref{fig:sr-explanation-b}c $\Delta_1 + \Delta_2 + \Delta_3 = 2$) towards computing to $w_3$, as the FP32 case, as depicted in Figure~\ref{fig:sr-explanation-b}b.

We now explain Theorem~\ref{thm:rounding} by considering a general case, where we round a gradient $x$ in a decision quantization interval $[a, b]$, as depicted in Figure~\ref{fig:sr-explanation-b}c. In this case, we express the weight gradient $x$ as $x = p(b-a)/q + a$ for some $p$ and $q$ with $0 \leq p < q$. (Note that if $a = 0$, $b = 1$, and $p = 2$, $q = 3$, then Figure~\ref{fig:sr-explanation-b}d depicts the scenario of Figure~\ref{fig:sr-explanation-b}a.)
Using SR, $x$ ($= p(b-a)/q + a$) is rounded to $a$ and $b$ with probability $(b-x)/(b-a)$ and $(x-a)/(b-a)$, respectively. Note that the two probabilities sum to 1, as expected. Thus, 
each iteration is expected to increment the weight value by
$a\times(b-x)/(b-a) + b\times(x-a)/(b-a)=x$,
which is the same weight increment under FP32 without rounding.

The $q$ value in Figure~\ref{fig:sr-explanation-b}c specifies the precision that SR uses itself. For example, for Figure~\ref{fig:fp-to-bfp}c, $q = 8$ since we add in 3 stochastic noise bits before rounding, and $2^3 = 8$.

\section{FAST Strategy for Training}
\label{sec:fast-training}

In this section, we describe our FAST strategy for DNN training which varies the BFP precision of weights, activations, and gradients based over the course of training. First, in Section~\ref{sec:fast-training:motivation}, we provide motivation for progressively increasing the precision across both training iterations and DNN layers. Then, in Section~\ref{sec:fast-training:alg}, we propose an approach to adaptively increase the precision during training.

\subsection{Progressive Precision Changes over DNN Training}
\label{sec:fast-training:motivation}
\begin{figure}
\centering
\includegraphics[width=\columnwidth]{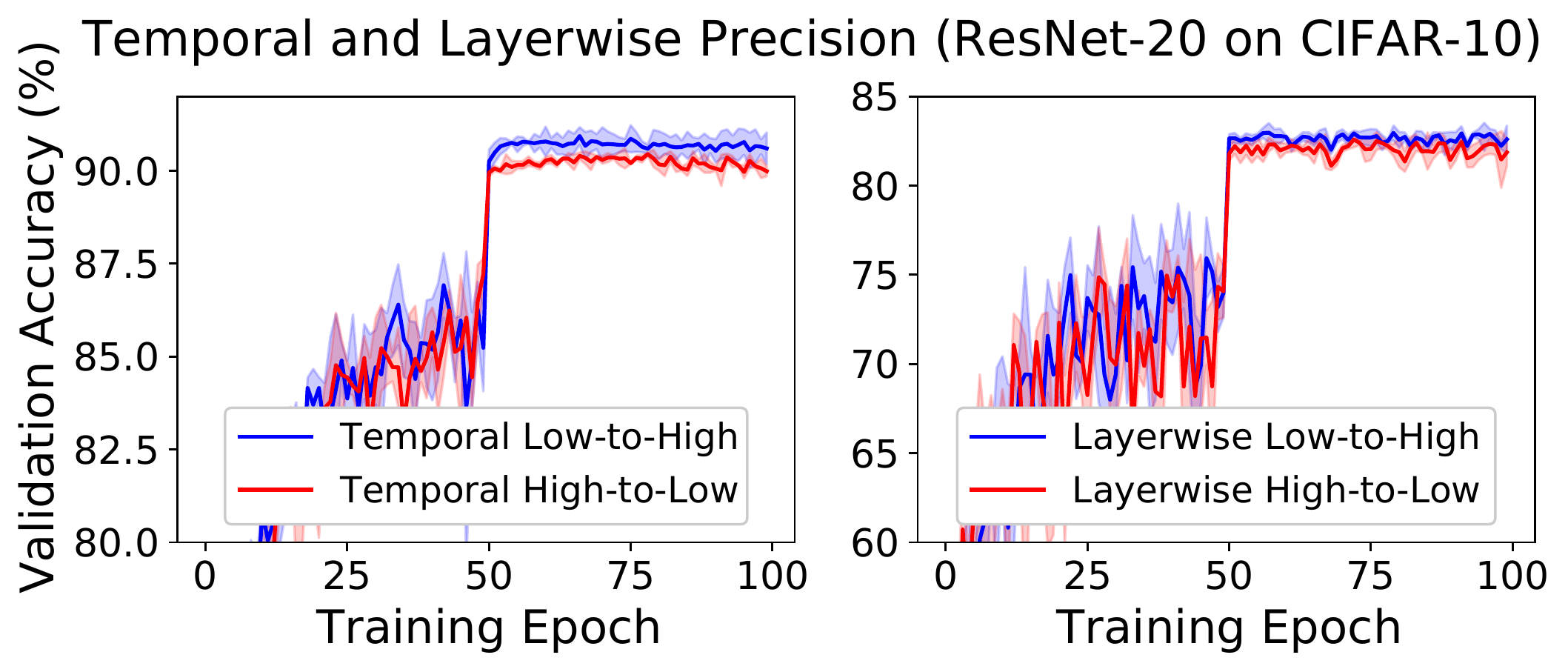}
\caption{(left) The Temporal Low-to-High scheme outperforms the Temporal High-to-Low scheme on validation accuracy. (right) The Layerwise Low-to-High scheme outperforms the Layerwise High-to-Low scheme. Each line is the mean of 3 runs, with the standard deviation shown as a shaded region.}
\label{fig:temporal-and-layer-precision}
\end{figure}

Previous literature has demonstrated that adding zero-mean Gaussian noise to the weight gradient $\nabla W$ can reduce overfitting and improve the convergence of DNN training~\cite{neelakantan2015adding}. They show that decreasing the variance of the noise over iterations achieves better performance than using fixed Gaussian noise throughout training. We hypothesis that a similar effect can be achieved by adjusting the BFP precision of weights, activations, and gradients from low to high precisions over training. To test this, we compare two training schemes (using ResNet-20 on CIFAR-10) that use different strategies for switching the DNN training precision over time. In the Temporal High-to-Low scheme, we use FP32 for weights, activations, and gradients for the first half of training, and low-precision BFP with a mantissa bitwidth of 3 and group size of 16 for the second half of training. For the Temporal Low-to-High scheme, we adopt the opposite approach by using low-precision BFP in the first half of training and FP32 in the second half of training. Figure~\ref{fig:temporal-and-layer-precision} (left) shows the test accuracy of these two schemes over the training process. The Low-to-High scheme achieves a higher performance, which indicates that training is more amenable to low-precision BFP in the early stages.

Additionally, during the backward pass of the training, the BFP quantization error for the data gradient $\nabla O$ will have a greater impact on the early layers than later layers. We perform another experiment to show this impact by comparing against two training schemes. In the Layerwise High-to-Low scheme, we use FP32 precision for the first ten layers, and low-precision BFP with a mantissa bitwidth of 3 and group size of 16 for later 10 layers. For the Layerwise Low-to-High scheme, we apply the opposite precision setting by switching the training precision between the first and second half of the DNN layers. To eliminate the impact on the architectural difference, we change the structure of ResNet-20 so that the first and the second halves have the same weight filter layout. The results shown in Figure~\ref{fig:temporal-and-layer-precision} (right) indicate that applying low precision in the early layers works better than later layers. 
\begin{algorithm}[t]
\caption{Adaptive FAST DNN Training}
\label{alg:adaptive-fast-training}
\small
\DontPrintSemicolon
  \KwIn{
  $I$ is the total number of training iterations. \newline
  $L$ is the total number of DNN layers. \newline
  $A_{l}$, $W_{l}$, $G_{l}$ are the activation, weight and gradient tensor of layer $l$, respectively. \newline
  $\epsilon(l,i)$ is a threshold to determine the BFP precision. \newline
  $BFP(X,m)$ is a BFP quantization function that returns X under BFP with an $m$-bit mantissa. 
  }
  \KwOut{$X_{q}$ represents the BFP-quantized X.}
  \For{$i \gets 1 \text{ to } I$}{
      \For{$l \gets 1 \text{ to } L$}{
          \For{$X\in [A_{l},W_{l},G_{l}]$ }{
              {Compute the relative improvement $r(X)$ for $X$.} \\
              \If{$r(X) < \epsilon(l, i)$}{
                  {Set $X_{q}$ = BFP(X, 2).} \\
              }
              \Else{
                  {Set $X_{q}$ = BFP(X, 4).}  \\
              }
          }
      }  
  }
\end{algorithm}
\subsection{FAST Adaptive Training}
\label{sec:fast-training:alg}
Based on the insight of the prior section, we propose an adaptive training strategy that progressively increase the BFP precision across both training iterations and layer depth. Algorithm~\ref{alg:adaptive-fast-training} describes the mechanism of the FAST training algorithm. FAST supports two precision levels by representing the BFP mantissas with either 2 bits (low precision) or 4 bits (high precision). For a given FP tensor $X$, FAST first evaluates the relative improvement $r(X)$ , defined by Equation~\ref{eqn:quant-error}, of using a 4-bit mantissa compared to a 2-bit BFP mantissa. If the relative improvement of using the higher precision setting is smaller than a threshold (i.e., $r(X)<\epsilon$), then a 4-bit mantissa does not offer significant improvement over a 2-bit mantissa. However, if the relative improvement is larger than the threshold, then using a 4-bit mantissa will significantly reduce the quantization error compared to a 2-bit mantissa.

To allow for an incrementally increasing BFP precision across both layer depth and training iterations, the threshold $\epsilon(l,i)$ is set to vary with the layer depth $l$ and training iteration $i$ based on the following equation:
\begin{align}
    \epsilon(l,i) = \alpha-\beta\frac{i}{I}-\beta\frac{l}{L} 
    \label{eqn:thres}
\end{align}
where $I$ and $L$ are the total number of training iterations and DNN layers, respectively. $\alpha$ and $\beta$ are the hyperparameters that specify the offset and the slope of the threshold function. Equation~\ref{eqn:thres} sets $\epsilon(l,i)$ to decrease gradually with both training iteration and layer depth, so that higher precisions will be used as the training iteration and layer depth grow. 
\begin{figure}[t]
\centering
\includegraphics[width=0.72\columnwidth]{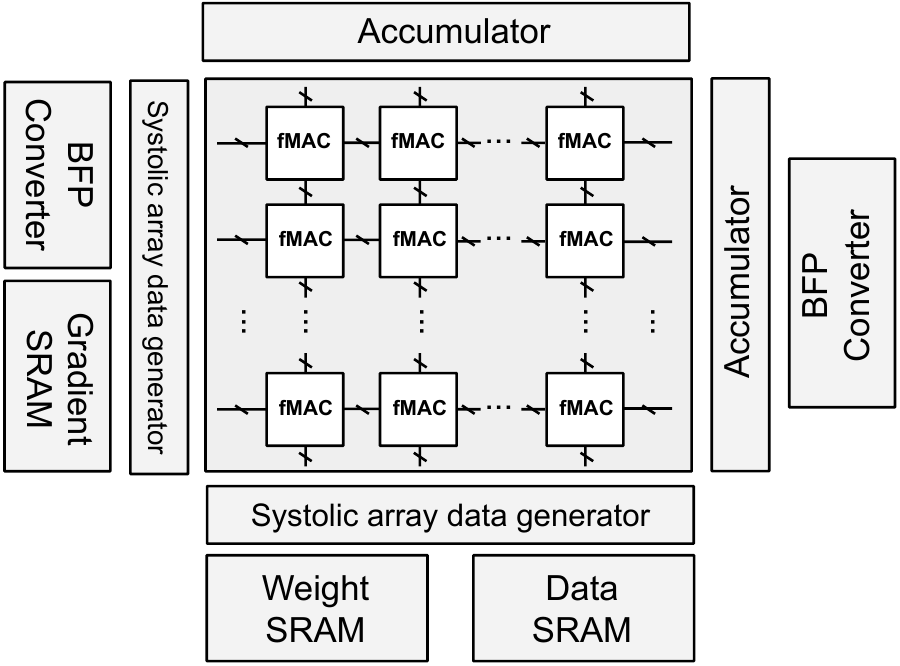}
\caption{Major components of the FAST system.}
\label{fig:systolic-system}
\end{figure}
We define the relative improvement $r(X)$ of using higher-precision BFP ($m=4$) compared to low-precision BFP ($m=2$) as follows:
\begin{equation}
    r(X) = \frac{\sum_{n}|BFP(X_{n},4)-BFP(X_{n},2)|}{\sum_{n}|BFP(X_{n},2)|}
    \label{eqn:quant-error}    
\end{equation}
where $X_{n}$ denotes the $n$th element of X, and $BFP(X, m)$ represents the quantized $X_{n}$ with an $m$-bit mantissa. The numerator of $r(X)$ reflects the total difference between the BFP values with high-precision and low-precision mantissas across each element of X. These difference is further divided by the summation of magnitudes of the BFP-quantized values so that the scale of $r(X)$ will be consistent across different training iteration and DNN layers. Finally, the numerator and denominator of $r(X)$ are computed by summing the BFP-quantized numbers across each element, which can be implemented with low hardware cost. The expensive division operation is only performed once between the two sums. 
\section{FAST System}
\label{sec:hw-arch}
The major components of the proposed FAST system are shown in Figure~\ref{fig:systolic-system}. We use a 2D systolic array (Section~\ref{sec:hw-arch:matrix-transpose}) to perform the matrix multiplications for both the forward and backward passes of DNN training. The systolic array contains systolic cells of FAST MAC (fMAC), discussed in Section~\ref{sec:hw-arch:fmac}, which support variable precision BFP. The memory subsystem has three SRAMs used to store weights, activations, and gradients, respectively. When performing matrix multiplication, the systolic array data generator is used to skew input for data synchronization in the systolic array. The accumulator is used to buffer partial accumulations across multiple tiles (for matrices that are larger than the systolic array). The output of the accumulator is passed to the BFP generator (Section~\ref{sec:hw-arch:bfp-generator}), which converts groups of FP values into BFP groups.

\subsection{Systolic Array Operations}
\label{sec:hw-arch:matrix-transpose}
\begin{figure}
\centering
\includegraphics[width=0.5\textwidth]{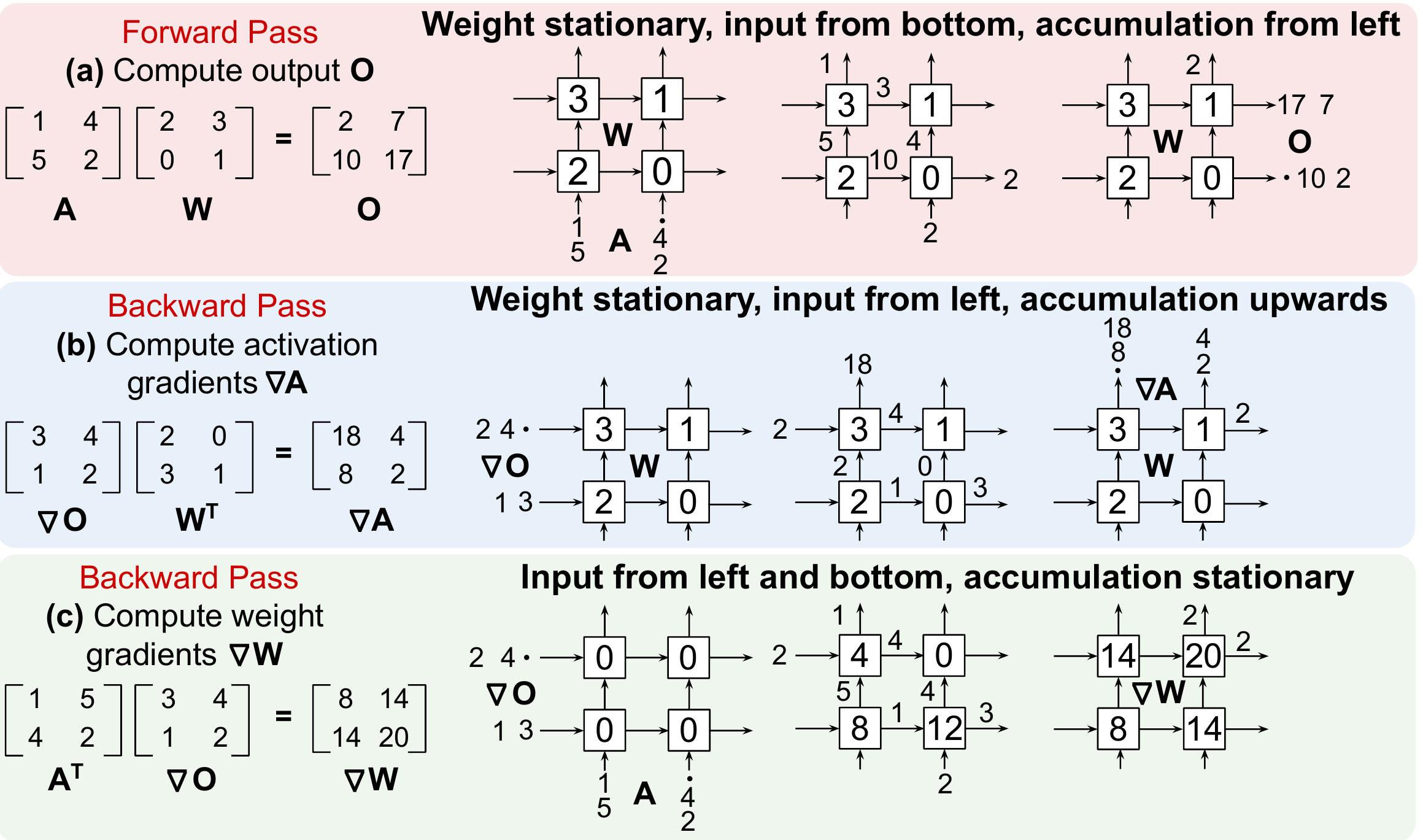}
\caption{Our systolic array can supports matrix multiplications required for both the forward and backward pass of DNN training. The matrix transposition required for matrix multiplications in the backward pass shown in (b) and (c) are handled by changing the side which the input enters the systolic array. Refer to Figure~\ref{fig:dnn-training-computation} for matrix notation.}
\label{fig:systolic-transpose}
\end{figure}

To support matrix transposition required during the backward pass of training (See Figure~\ref{fig:dnn-training-computation}), we have developed a systolic array that can perform matrix multiplication involving a transposed matrix operand without explicit transposition. This allows for no extra data copying and thus reduces the implementation overhead of the matrix transposition operation. In Figure~\ref{fig:systolic-transpose}, we illustrate how this systolic array operates for each of the forward and backward pass matrix operations given in Figure~\ref{fig:dnn-training-computation}. For clarity, we show each systolic cell with single INT values instead of a BFP group.

To compute output $O$ for each layer (Figure~\ref{fig:systolic-transpose}a), the weights $W$ are first pre-stored in systolic cells. Then, the activation $A$ enters the systolic array from bottom and the output $O$ exits the systolic array from the right side (refer to Figure~\ref{fig:dnn-training-computation}). During the backward pass, to compute the activation gradients $\nabla A$ (Figure~\ref{fig:systolic-transpose}b), $W$ is also pre-stored in the systolic array. However, unlike the forward pass, with $A$ entering from below, the output gradients $\nabla O$ enter the systolic array from left and the activation gradients $\nabla A$ are produced at the top of systolic array. By changing the side that input enters the systolic array while keeping the orientation of $W$ fixed, we can compute $\nabla O \times W^T = \nabla A$ without explicitly transposing $W$. 

\begin{figure}
\centering
\includegraphics[width=0.85\columnwidth]{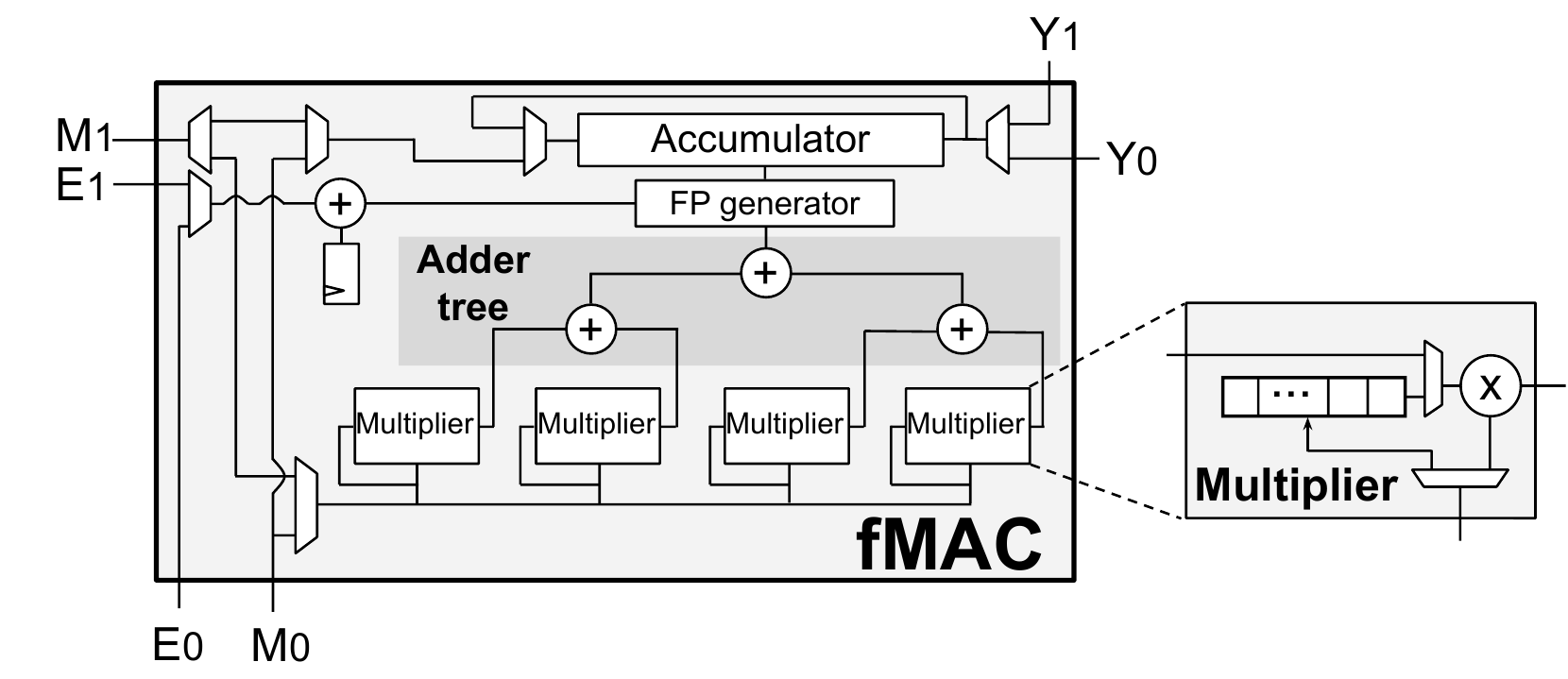}
\caption{The design of FAST MAC (fMAC).}
\label{fig:fmac}
\end{figure}

Finally, to compute the weight gradients $\nabla W$ (Figure~\ref{fig:systolic-transpose}c), the systolic array is reconfigured to be accumulation stationary. During the computation, the input activation $A$ and output gradient $\nabla O$ will enter the systolic array from left and below, respectively, and the weight gradients $\nabla W$ are computed and accumulated within each systolic cell.  At the end of this computation, the accumulated gradient in each systolic cell will sum with the weight $W$ to generate the updated weight $W'$, which is then stored back in the weight SRAM. For optimizer like Adam~\cite{kingma2014adam}, additional hardware is required to compute the first and second moments for the weight updates.

\subsection{Design of FAST MAC for BFP}
\label{sec:hw-arch:fmac}
Each cell in the systolic array implements a fMAC, which perform the DP between two BFP groups. Figure~\ref{fig:fmac} shows the design of a fMAC for a group size $g=4$. A DP consists of fixed-point multiplications between each pair of BFP mantissas for values in the groups, which are performed by the multipliers in the fMAC. The output generated by each multiplier is summed using the adder tree. The FP generator takes the fixed point summation from the adder tree to create the FP mantissa. The shared exponents of the two BFP groups are added together to create the FP exponent (refer to Figure~\ref{fig:bfp-dot-product}). The resulting FP value is adding to the FP accumulator which stores the partial result spanning across many BFP groups.

Additionally, the fMAC can be reconfigured to support the different operations of DNN training described in Section~\ref{sec:hw-arch:matrix-transpose} which may require matrix transposition. To compute the output $O$ during the forward pass (Figure~\ref{fig:systolic-transpose}a), the BFP shared exponent and mantissas of $W$ are first pre-stored in the fMAC using the $E_{0}$ and $M_{0}$ ports, respectively. Then, the activation $A$ enter the fMAC via the same $E_{0}$ and $M_{0}$ ports to perform the DP computation. The output $O$ generated by this computation exits to the right neighbor via output port $Y_{0}$. To compute the activation gradient $\nabla A$ during the backward pass (Figure~\ref{fig:systolic-transpose}b), $W$ is pre-stored in the same fashion, and the BFP output gradients $\nabla O$ are passed into the multipliers via the $E_{1}$ and $M_{1}$ input ports, with the output $\nabla A$ exiting to the neighbor  above via $Y_{1}$. Finally, to compute the weight gradients $\nabla W$ (Figure~\ref{fig:systolic-transpose}c), the output gradients $\nabla O$ and input activation $A$ enter the fMAC using the ports $M_{1}$, $E_{1}$ and $M_{0}$, $E_{0}$, respectively. The accumulator output $\nabla W$ then loops back to be summed with the pre-stored FP weights $W$.

\begin{figure}
\centering
\includegraphics[width=0.9\columnwidth]{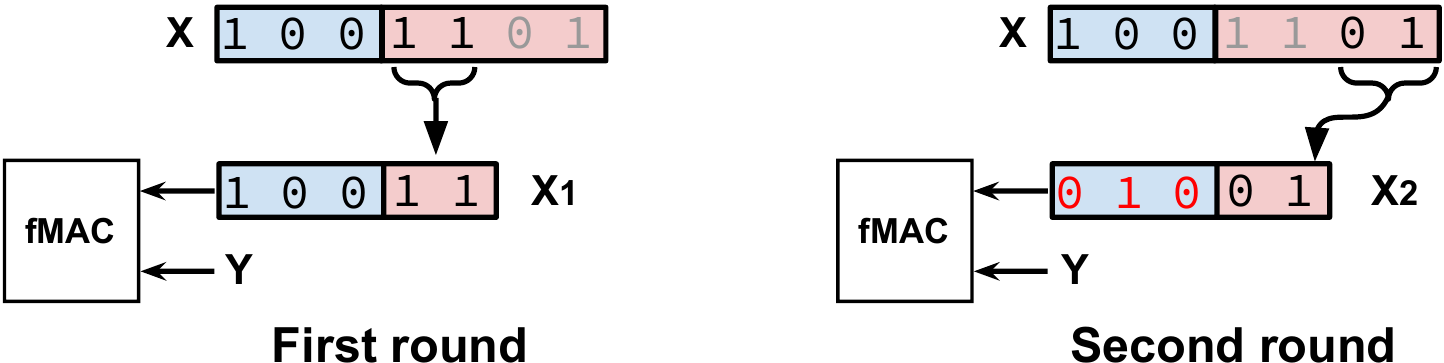}
\caption{Variable precision fMAC operations.}
\label{fig:variable-precision-computing}
\end{figure}

To support BFP DP with variable precision, each DP is processed in 2-bit mantissa chunks as shown in Figure~\ref{fig:variable-precision-computing}. Here, the mantissas bitwidths for two operands ($X$ and $Y$) are 4 bits and 2 bits, respectively. In the first round, the fMAC computes the dot product between Y and the first 2-bit chunk of the X ($X_{1}$). The partial accumulation result is then buffered for subsequent processing. In the second round, the fMAC computes the dot product between $Y$ and the second 2-bit chunk of X ($X_{2}$) in order to finish the DP computation. More iterations are required for higher mantissa bitwidth. For example, multiplying a pair of BFP numbers with 4-bit and 4-bit mantissas translates to $\frac{4}{2}\times \frac{4}{2} = 4$ rounds. To account for the difference in exponent magnitude between two chunks, the BFP exponent of the second 2-bit chunk ($X_{2}$) is decremented by two. Note that this decrement is performed by the BFP converter when it generates each 2-bit chunk, and therefore fMAC is agnostic to these exponent difference across chunks.

\subsection{BFP Converter}
\label{sec:hw-arch:bfp-generator}
\begin{figure}
\centering
\includegraphics[width=0.9\columnwidth]{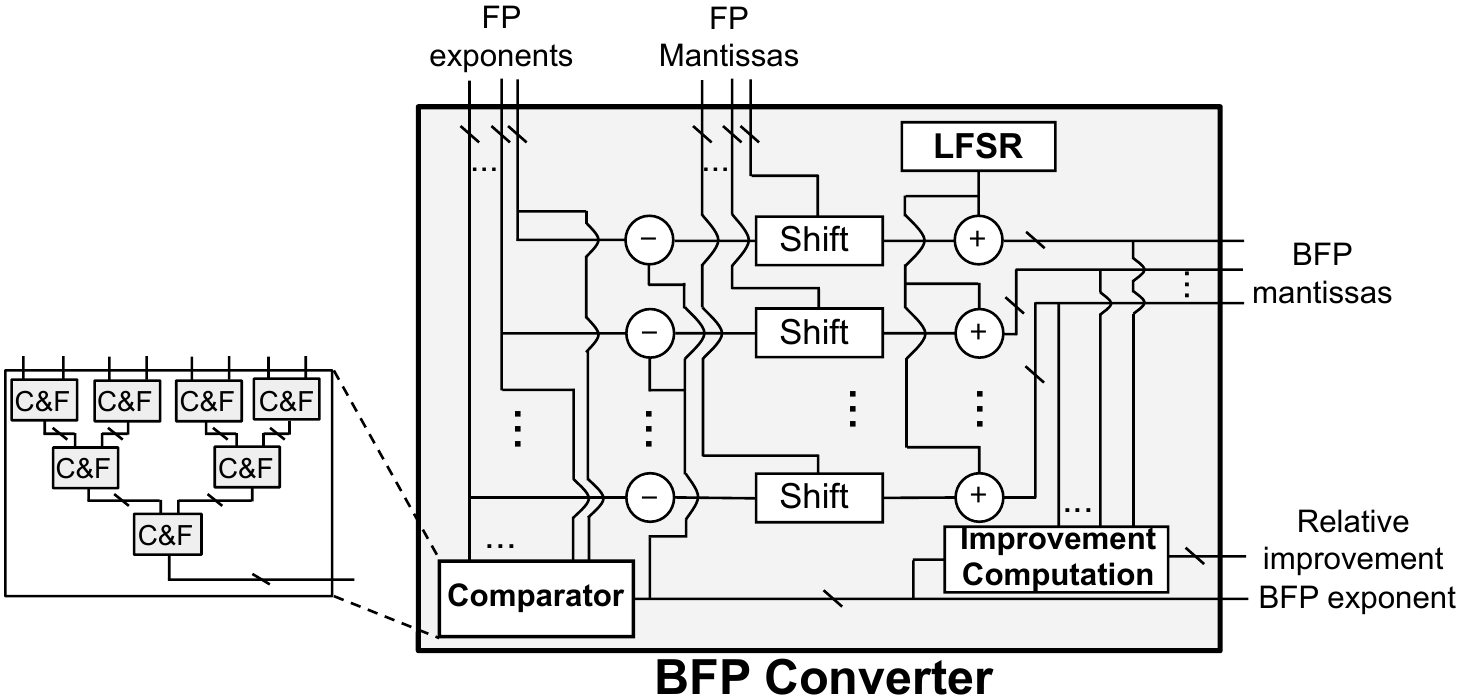}
\caption{Design of BFP converter which converts a group of FP values into BFP values.}
\label{fig:fp-tp-bfp-converter}
\end{figure}

The BFP converter, shown in Figure~\ref{fig:fp-tp-bfp-converter}, takes a group of FP values and converts them into BFP following the process outlined earlier in Figure~\ref{fig:fp-to-bfp}. The comparator consists of compare and forward (C$\&$F) blocks arranged in a tree structure. Each C$\&$F block takes a pair of FP exponents and forwards the larger exponent to the next tree level. The largest exponent will be output and used as the shared exponent (Figure~\ref{fig:fp-to-bfp}a). Then, a group of subtractors calculate the differences between the shared exponent and each exponent in the group. The shift blocks, which are implemented using Barrel shifters~\cite{pillmeier2002design}, perform right shifts on each FP mantissa based on the exponent difference for each value (Figure~\ref{fig:fp-to-bfp}b). Then, to perform stochastic rounding (Figure~\ref{fig:fp-to-bfp}c), a group of 8-bit random binary streams produced by the linear feedback shift register (LFSR) are summed with the mantissas. Finally, the low-order bits of the BFP mantissas are truncated (Figure~\ref{fig:fp-to-bfp}d). The BFP exponents and BFP mantissas will also be delivered to the improvement computation block which computes the relative improvement as defined in Equation~\ref{eqn:quant-error}.

\subsection{Memory Layout for BFP Values}
\label{sec:hw-arch:memory-layout}
\begin{figure}
\centering
\includegraphics[width=0.9\columnwidth]{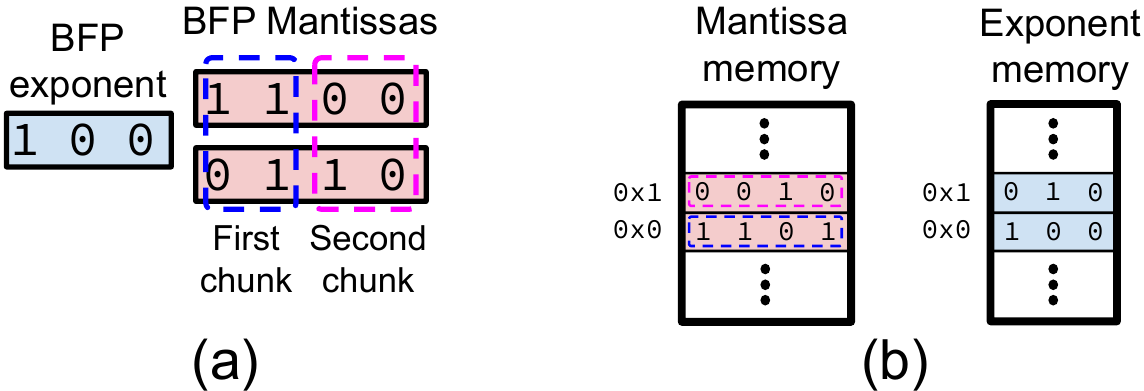}
\caption{The BFP memory layout for $g = 2$ and $m$ = 4. Each mantissa is divided into 2-bit chunks to support variable precision multiplications. Sign bits are not shown.}
\label{fig:memory-layout}
\end{figure}

We have developed an efficient storage format for variable precision BFP, where the shared exponent and BFP mantissas are stored separately. Figure~\ref{fig:memory-layout} provides an example for $m$ = 4 and $g$ = 2. The 2-bit chunks across all the mantissas in a group are saved in the same memory entry for efficient access during DP computation. The first 2-bit chunks of each BFP mantissa (Figure~\ref{fig:memory-layout}a) are stored together in the same memory entry (Figure~\ref{fig:memory-layout}b), followed by the second 2-bit chunks which are saved in the next memory entry. 

Under this storage scheme, each BFP group will be represented by $e+g\times \frac{m}{2}\times 3$ bits, where $e$ is the bitwidth of the BFP exponent, $g$ is the group size, $\frac{m}{2}$ is the number of 2-bit chunks in an m-bit mantissa. An additional bit is required per mantissa to represent the sign, leading to 3 bits per mantissa. In our hardware system, $e$ and $g$ are set to be 3 and 16, respectively, and $m$ is $2$ or $4$ based on the current precision. This leads to an average of $3.2$ ($m=2$) and $6.2$ ($m=4$) bits to store each value, which significantly reduces the storage overhead compared with other formats we evaluate in Section~\ref{sec:sw-eval}.  
All outputs from the BFP converter (Figure~\ref{fig:fp-tp-bfp-converter}) are stored in BFP with 4-bit mantissas divided into two 2-bit chunks. If Algorithm~\ref{alg:adaptive-fast-training} selects the 2-bit mantissa, then the low-order 2-bit chunk is discarded.

\subsection{Training Workflow}
\begin{figure}
\centering
\includegraphics[width=\columnwidth]{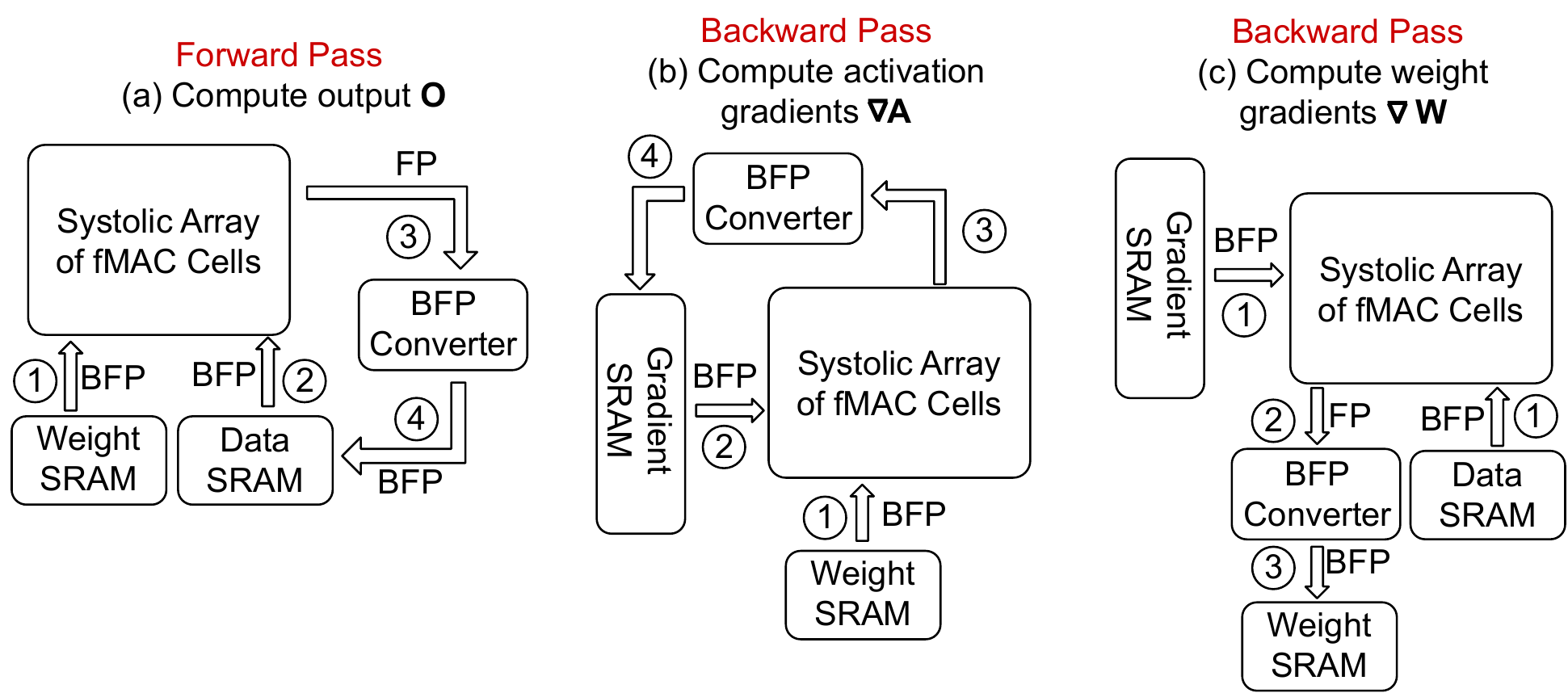}
\caption{Workflow during each training iteration, where the systolic array operates according to Figure~\ref{fig:systolic-transpose}.}
\label{fig:dnn-training-systolic}
\end{figure}

Figure~\ref{fig:dnn-training-systolic} summarizes the overall workflow of DNN training using the FAST system. During the forward pass (Figure~\ref{fig:dnn-training-systolic}a), the filter weights $W$ in BFP format are first loaded and saved into each fMAC (step 1). Next, the BFP activations $A$ are loaded from data SRAM into the systolic array (step 2). Each systolic cell (fMAC) performs a partial DP for two BFP groups, followed by FP accumulation spanning across many BFP groups. The output $O$ is then delivered to the BFP converter (step 3), which converts these values back into BFP format and stores them in the data SRAM for subsequent processing (step 4). Additionally, the activations $A$ must also be kept in the data SRAM for the backward pass (Figure~\ref{fig:dnn-training-systolic}b-c).

\begin{figure*}
\begin{minipage}{0.68\textwidth}
\vspace{18pt}
\centering
\captionsetup{type=table} 
        
\begin{adjustbox}{width=1\columnwidth,center}
\begin{tabular}{|c|c|c|c|c|c|c|c|c|c|c|c|}
\hline
Model & FP32  & bfloat16 & Nvidia MP  & INT8  & INT12 & MSFP-12 & LowBFP & MidBFP & HighBFP & {HFP8} &  FAST \\ \hline
ResNet-18    & 68.60 & 68.55  & 68.57 & 65.53 & 68.51 & 68.13   & 63.10  & 68.10 & 68.57 & {68.53} & 68.52  \\ \hline
ResNet-50    & 75.17 & 75.12 & 75.16  & 71.01 & 75.03 & 74.79   & 72.10  & 73.98 & 75.13 & {75.07}  &  75.11  \\ \hline
MobileNet-v2 & 68.27 & 68.22  & 68.28 & 65.97 & 68.16 & 68.11   & 64.42  & 66.93 & 68.20 & {68.11} &  68.17  \\ \hline
VGG-16       & 69.74 & 69.71 & 69.70 & 64.50 & 69.33 & 69.32   & 64.10  & 69.08 & 69.79 & {69.62} &  69.78  \\ \hline
Transformer  & 35.41 & 35.39 & 35.42 & 29.18 & 35.27 & 35.33   & 34.22  & 35.40 & 35.43 & {35.38} & 35.40   \\ \hline
{YOLOv2}  & {73.36} & {73.32} & {73.35} & {61.12} & {73.07} & {72.93}   & {65.37} & {71.04} & {73.30}  & {72.88} & {73.28}   \\ \hline
\end{tabular}
\end{adjustbox}
\caption{The validation accuracy (CNNs), test BLEU (Transformers) and test mAP (YOLOv2) for number formats outlined in Figure~\ref{fig:fp-formats}.}
\label{tab:arith-acc}
\end{minipage}
\begin{minipage}{0.3\textwidth}
\vspace{18pt}
\captionsetup{type=table} 
        \centering
        \begin{adjustbox}{width=0.98\columnwidth,center}
\begin{tabular}{l|c|c}
Component  & Area & Power \\ \hline
Systolic array    &     47.79\%   &       15.61 W    \\
BFP converter  &    4.56\%   &     1.77 W  \\
Accumulator &    6.63\%    &      2.19 W \\
Systolic array data generator  &    0.68\%    &     0.69 W  \\
Memory subsystem    &     40.34\%   &       3.37 W    \\
\end{tabular}
        \end{adjustbox}
\caption{Area and power breakdown of the FAST system.}
\label{tab:area-power-breakdown}
\end{minipage}

\end{figure*}

To compute the activation gradients $\nabla A$ (Figure~\ref{fig:dnn-training-systolic}b), the weights $W$ are again pre-stored into systolic array (step 1). Then, the output gradients $\nabla O$ are delivered to the systolic array from the left (step 2). The results $\nabla A$ are produced at the top of systolic array and are converted into BFP format (step 3) before being saved into the gradient SRAM (step 4). To compute $\nabla W$ (Figure~\ref{fig:dnn-training-systolic}c), the input activation $A$ and the output gradient $\nabla O$ are delivered to the systolic array concurrently (step 2). The results $\nabla W$ are produced within each systolic cell and then they are used to generate the updated weight $W'$. Finally, the updated weights $W'$ are converted into BFP and stored in weight SRAM (step 3). 

\section{Training Evaluation of FAST}
\label{sec:sw-eval}
In this section, we evaluate FAST's training performance for DNNs. In Section~\ref{sec:sw-eval:heatmap}, we visualize the FAST precision adaptation over the course of training to show how FAST is able to achieve faster training time by staying in a low-resolution regime for a large portion of training. Next, in section~\ref{sec:sw-eval:float-comp}, we compare the accuracy performances of DNNs trained under BFP against other commonly used FP and INT formats. We also compare against three fixed BFP settings that do not change over the course of training: LowBFP uses ($e$=3, $m$=2) for all DNN weights, data, and activations, MidBFP uses ($e$=3, $m$=3), and HighBFP uses ($e$=3, $m$=4). Finally, in Section~\ref{sec:sw-eval:bfp-params}, we evaluate the performance of fixed BFP settings to show the relative advantage of using FAST. All the CNNs are trained with ImageNet for 60 epochs (120000 iterations). We use the hyperparameter settings from the PyTorch website~\cite{pytorch-settings}.

For Transformers, we use the 12-layer model with 12 heads and a hidden size of 768. The Transformer is trained using the Adam optimizer with a learning rate of $10^{-4}$, $\beta_{1}=0.9$ and $\beta_{2}=0.999$. The batch size is set to 16. We train on the IWSLT14 German-English dataset~\cite{iwslt-benchmark} for 150 epochs. Finally, we train YOLOv2~\cite{redmon2017yolo9000} on the PASCAL VOC2012~\cite{everingham2011pascal} dataset with 120 epochs using a batch size of 64. We apply the SGD optimizer with a initial learning rate of $10^{-3}$, dividing it by 10 at 60 and 90 epochs, the weight decay and momentum are set to 0.0005 and 0.9. The $\alpha$ and $\beta$ FAST hyperparameters in Equation~\ref{eqn:thres} are set to 0.6 and 0.3 for all the DNNs.

\subsection{FAST Precision Adaptation}
\label{sec:sw-eval:heatmap}
\begin{figure*}
\centering
\begin{minipage}[t]{0.42\textwidth}
  \vspace{0pt}
  \centering
  \includegraphics[width=\columnwidth]{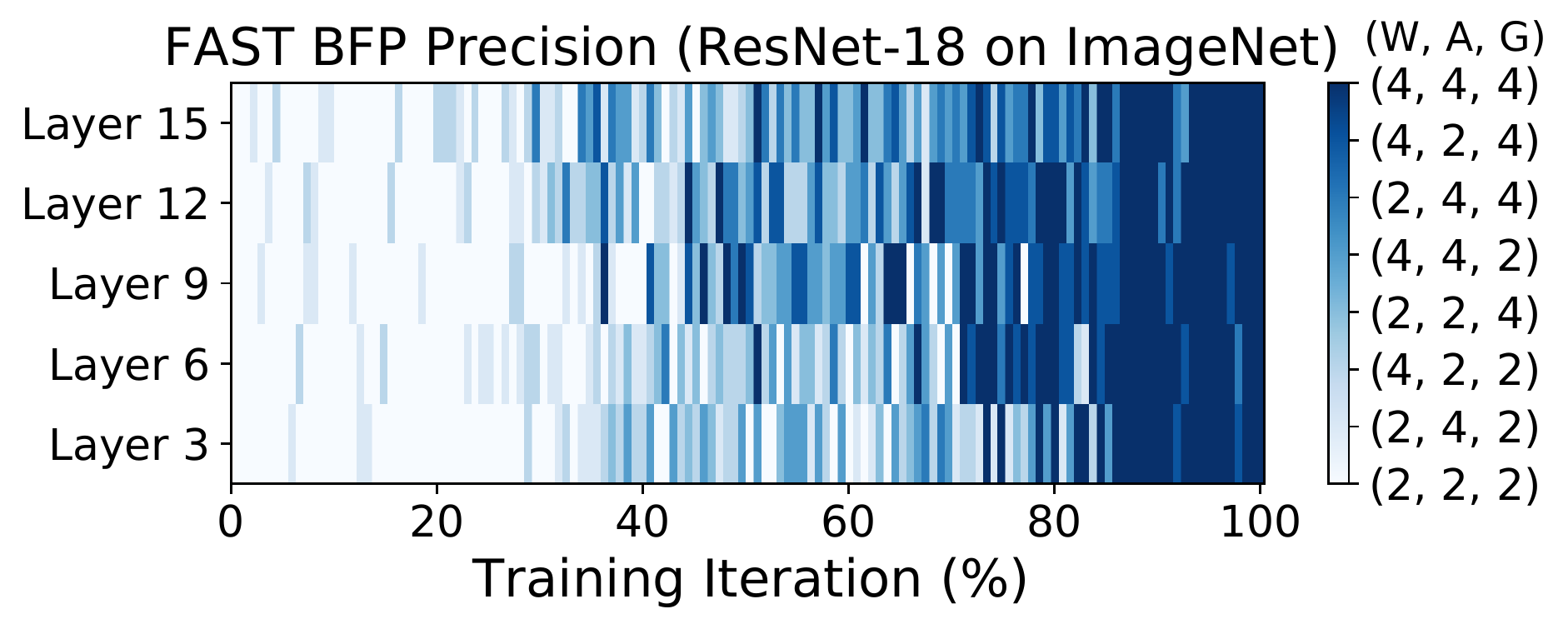}
  \caption{FAST progressively increases the BFP precision across both layer depth and iterations during the training process.}
  \label{fig:heatmap}
\end{minipage}\hfill%
\begin{minipage}[t]{0.23\textwidth}
  \vspace{0pt}
  \centering
  \includegraphics[width=0.97\columnwidth]{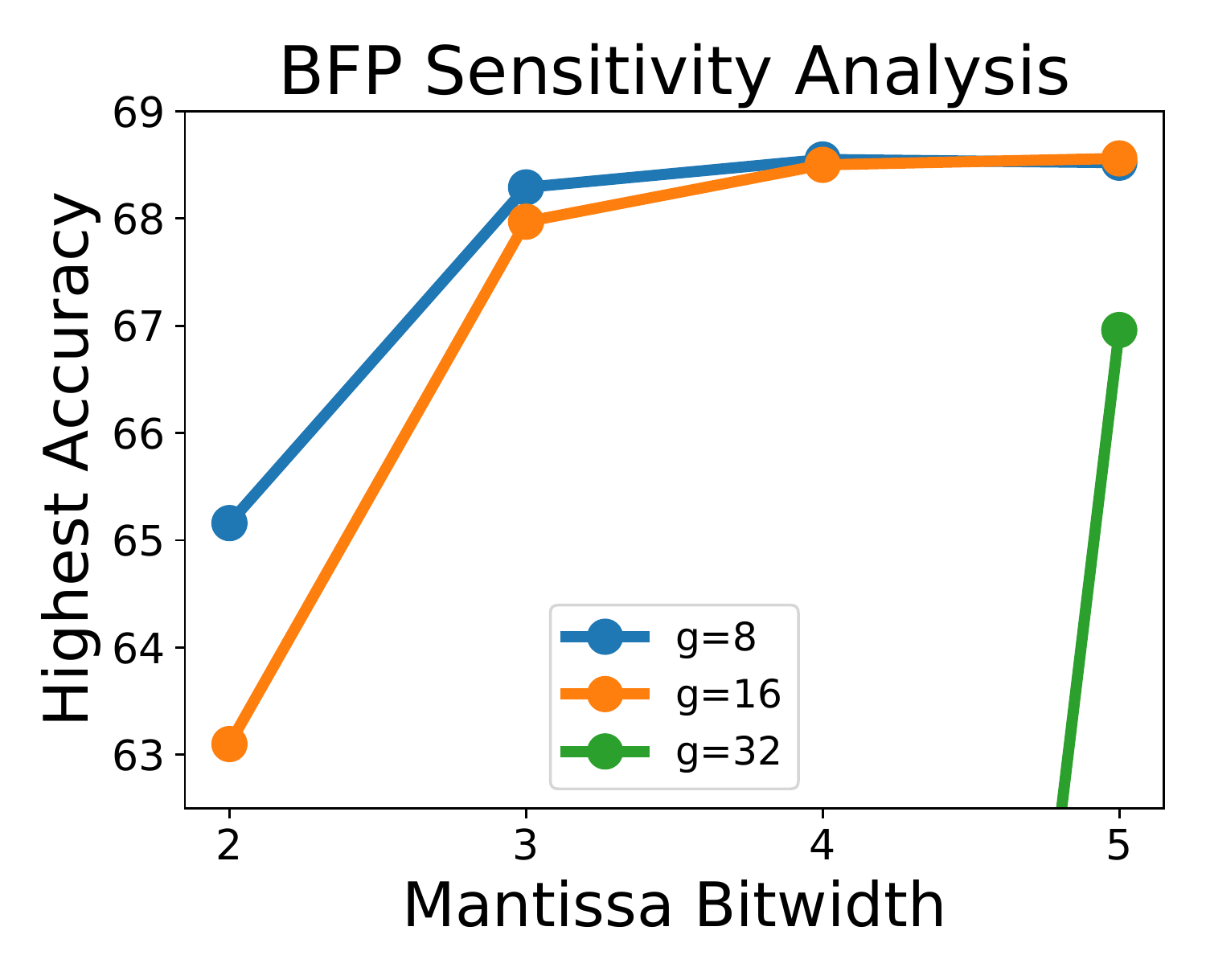}
  \caption{BFP sensitivity analysis (ResNet-18 on ImageNet).}
  \label{fig:blocksize-mantissa}
\end{minipage}\hfill%
\begin{minipage}[t]{0.32\textwidth}
  \vspace{0pt}
  \centering
  \includegraphics[width=\columnwidth]{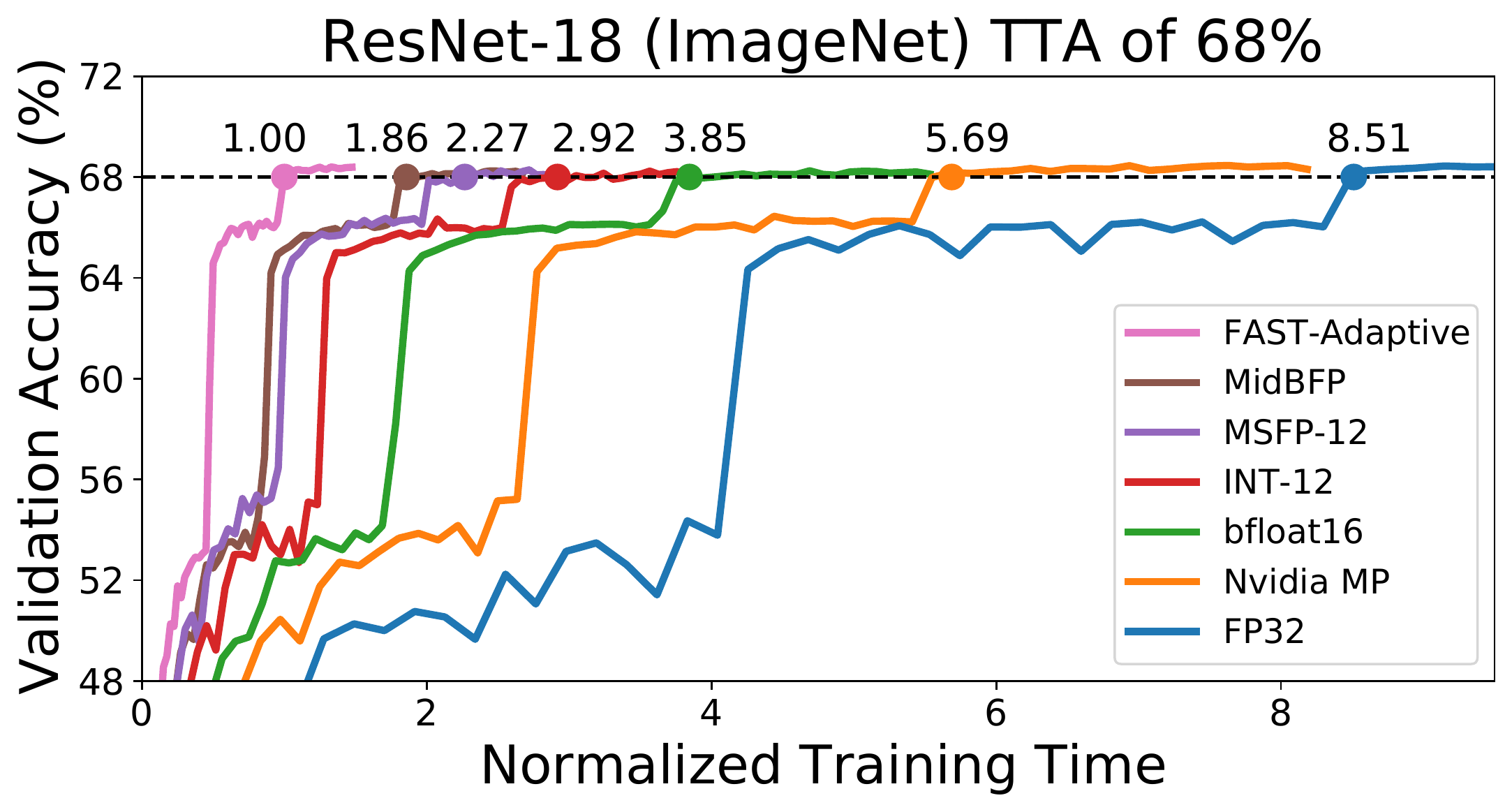}
   \caption{The validation accuracy curves and TTA values for achieving the 68\% accuracy for ImageNet (dotted line).}
   \label{fig:tta-curves}
\end{minipage}%
\label{fig:acc-figs}
\end{figure*}

In this section, we visualize the precision changes during the training of a DNN using the FAST-Adaptive algorithm (Algorithm~\ref{alg:adaptive-fast-training}). 
Since the weights $W$, activations $A$, and gradients $G$ independently determine their BFP precision, there are $2^3 = 8$ possible precision settings per layer using two different BFP resolutions ($m=2$ and $m=4$). In the figure, we have ordered these settings based on their computational costs when deployed in the FAST system (discussed next in Section~\ref{sec:hw-eval}). For instance, ($W$, $A$, $G$) of ($4$, $2$, $2$) has a slightly lower computational cost than ($2$, $2$, $4$) due to how the gradients are used multiple times during the backward pass (see Figure~\ref{fig:dnn-training-computation}). Figure~\ref{fig:heatmap} shows the BFP precisions of 5 layers in ResNet-18 on ImageNet change over the course of training under FAST. As expected, we observe that the BFP precision grows across both layer depths and training iterations.

\subsection{Comparing Number Formats}
\label{sec:sw-eval:float-comp}

Table~\ref{tab:arith-acc} shows the validation accuracies for different DNN models trained using a wide range of number formats. The IEEE 754 32-bit FP (FP32) generally achieves the best performance (accuracy or BLEU) across all models. We find that bfloat16, Nvidia Mixed-Precision (MP), MSFP-12 and HFP8 are able to achieve similar performance as the baseline FP32 model for all DNNs. Additionally, the HighBFP setting is also able to achieve comparable accuracy. HighBFP with $m$ = 4 represents a substantial saving compared to FP32 with $m$ = 23. The LowBPF and MidBFP settings with $m$ = 2 and $m$ = 3 loss $4-5\%$ and $1-2\%$ in accuracy across all CNNs compared to the FP32 baseline. The INT-8 setting has an even larger reduction in accuracy, losing 4-6\% compared to the baseline, even though it has more mantissa bits over HighBFP. For fixed point to achieve a similar level of performance as the baseline FP model requires an \textbf{INT-12 with 11 mantissa bits}. As we note in Section~\ref{sec:hw-eval}, fixed point multipliers used to perform this computation incur cost quadratically with the mantissa bitwidth, making large mantissa bitwidths costly to implement. By comparison, our FAST-Adaptive approach can achieve a comparable performance to FP32 across all the DNNs.

\subsection{BFP Hyperparameter Sensitivity}
\label{sec:sw-eval:bfp-params}

In this section, we investigate the impact of the group size $g$ and mantissa bitwidth $m$ on the DNN training accuracy. Figure~\ref{fig:blocksize-mantissa} shows the validation accuracy on ResNet-18 for different BFP configurations settings. The three curves represent different group size configurations (i.e., $g=8$, $g=16$, and $g=32$) and the x-axis corresponds to varying the number of mantissa bits (i.e., $m=2$, $m=3$, $m=4$ and $m=5$) for each group size.  For a given mantissa bitwidth, a smaller group size (e.g., $g=8$) is generally able to achieve a higher accuracy than a larger group size (e.g., $g=32$). However, a smaller group size has some additional implementation overhead due to more FP exponent additions as each shared exponent spans fewer elements in a smaller group. Overall, we observe that a group size of $g=16$ with $m=4$ produce the optimal performance, and we use this setting as our baseline for FAST training.

\section{Hardware Evaluation of FAST}
\label{sec:hw-eval}
In this section, we evaluate the hardware performance of the FAST system described in Section~\ref{sec:hw-arch}. We have synthesized our system using the Synopsys Design Compiler~\cite{synopsysdesigncompiler} with 45nm NanGate Open Cell Library~\cite{nangatelib} and CACTI~\cite{cacti}. CACTI is used to simulate the performance of the memory subsystem and Synopsys Design Compiler is used for all other subsystems shown in Figure~\ref{fig:systolic-system}. For our FPGA evaluation, we use a Xilinx VC707 FPGA evaluation board. The FAST system contains a $256\times 64$ systolic array of fMAC cells. Gradient SRAM, weight SRAM and data SRAM each consist of 128 16kB memory banks. The FAST system runs at 500 MHZ. Table~\ref{tab:area-power-breakdown} summarizes the area and power breakdowns of FAST. 
\begin{figure*}[h]
    \centering
    \includegraphics[width=\textwidth]{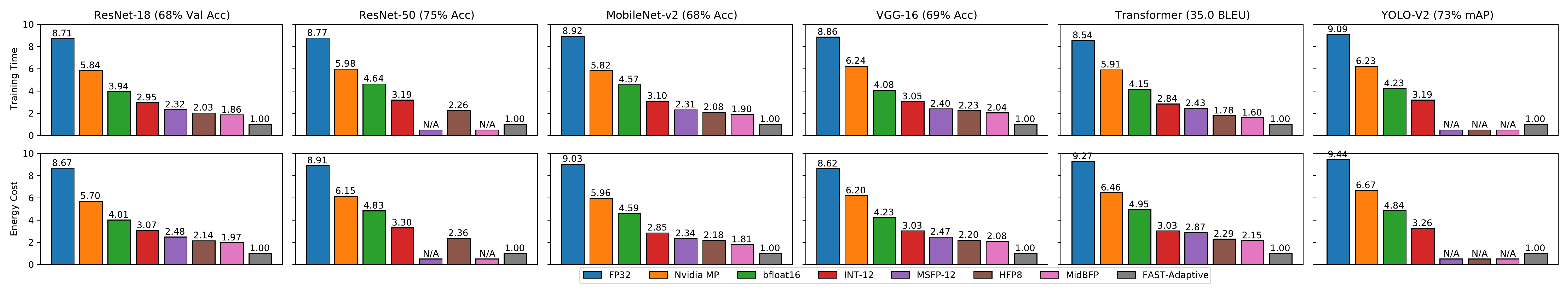}
    \caption{The normalized training time and energy cost comparisons under different number formats, which are color coded as in Figure~\ref{fig:tta-curves}. N/A means the target accuracy is never reached for a given setting.}
    \label{fig:asic-results}
\end{figure*}
\subsection{Evaluation of fMAC}
\label{sec:hw-eval:mac-comp}
We evaluate the efficiency of our fMAC design by comparing it against FP and INT MAC designs. For FP MACs, we implement them with bfloat16, FP16 and HFP8. FP16 is used by Nvidia MP. An FP MAC performs multiply-accumulate operations between two FP numbers followed by a 32-bit FP accumulation. For INT MACs, we implement them with 8-bit (INT-8) and 12-bit (INT-12) variants. Refer to Figure~\ref{fig:fp-formats} for details on each number format. Two floating point formats are used by HFP8 during training: 4-bit exponent/3-bit mantissa for the forward pass and 5-bit exponent/2-bit mantissa for the backward pass. For a hardware cost comparison to FAST, we implement a MAC that supports a 4-bit exponent/2-bit mantissa, so that the hardware cost is strictly less than either floating point format used by HFP8. Since a single fMAC performs a BFP DP across two groups of $g=16$ numbers, we use $g=16$ for all other MAC designs for a fair comparison.

Table~\ref{tab:fmac-evaluation} provides an ASIC evaluations in terms of area and power consumption and FPGA resource consumption for all MAC designs. Area consumption is normalized by the area of our fMAC design. The fMAC achieves a superior area and power consumption compared to the other MAC designs. The main advantage of the fMAC over the INT MAC designs is the significantly reduced mantissa bitwidth, leading to a substantial reduction in cost of the fixed point multipliers. When comparing fMAC to the FP MACs, the expensive FP accumulator is amortized over the group for fMAC instead of between each pair of elements for FP MACs. 

\subsection{Training Speedup of FAST Strategies}
\label{sec:hw-eval:dynamic}

In this section, we evaluate the performance of the FAST system by comparing against the other DNN training systems implemented with systolic arrays using different number formats. We configure each system for a given number format to have the same total area as our FAST system. Specifically, with the same area, we are able to fit a DNN training system with a systolic array of $245 \times 245$ HFP8 (4-bit exponent and 2-bit mantissa) MACs, $230 \times 230$ MSFP-12 MACs, $210 \times 210$ INT-12 MACs, $180 \times 180$ bfloat16 MACs and $150 \times 150$ FP16 MACs, respectively. Note that our FAST system contains a $256 \times 64$ fMAC systolic array, and each fMAC can perform multiply-accumulate operations for 16 BFP numbers within one cycle. The design for other major components (i.e., accumulator, numerical converter, systolic array data generator and memory subsystem) of the baseline DNN systems are modified according to a given MAC design. For example, for bfloat16, a bfloat16 converter is used instead of a BFP converter. All designs run at a 500MHz clock frequency.

We use Time-to-Accuracy (TTA)~\cite{coleman2017dawnbench} as the evaluation metric to compare different approaches. Figure~\ref{fig:tta-curves} shows the TTA for ResNet-18 models trained under various number formats to achieve a validation accuracy of 68\% on ImageNet. The training time is normalized by the FAST-Adaptive model which achieves 68\% the fastest. Some settings that were unable to achieve 68\% validation accuracy, such as INT8 and LowBFP, were omitted. The results are measured by performing a single round of forward pass and backward pass with a input mini-batch of size 256. The evaluation results are generated based on the computation required for all convolutional and fully connected layers. Normalization and Activation layers are not considered in the cost analysis of this paper. Prior work suggests that activations and batch normalization take less than 5\% of total running time~\cite{fleischer18}, and a small amount of power relative to the systolic array and other components~\cite{lee2020,qin2020}.


\begin{table}
\centering
\caption{ASIC area and power comparison and FPGA resource consumption for different MAC designs.}
\begin{adjustbox}{width=0.8\columnwidth,center,}
\begin{tabular}{l?c|c?c|c}
  & \multicolumn{2}{c?}{ASIC} & \multicolumn{2}{c}{FPGA}\\ \cline{2-5}
MAC Design & Area & Power & LUT & FF\\ \hline
fMAC    &    $1\times$    &      0.885mW  &    269    &      140   \\
16$\times$ INT-8 &    $3.8\times$    &      2.241mW &   498   &      195  \\
{16$\times$ HFP8}  &    {$4.1\times$}    &      {2.406mW} &   {527}    &   {220} \\
16$\times$ INT-12 &    $5.6\times$    &      2.920mW &   730    &     273 \\
16$\times$ bfloat16  &    $9.6\times$    &      3.869mW  &    1305    &      684 \\
16$\times$ FP16  &   $10.6\times$    &      4.474mW &    1514    &  753\\
\end{tabular}
\end{adjustbox}

\label{tab:fmac-evaluation}
\end{table}
Generally, we see that FP32 is significantly slower than reduced/mixed precision formats such as bfloat16 and Nvidia MP. However, the floating point accumulations required for each MAC using these formats introduces overhead compared to fixed point of BFP formats. The MSFP-12 achieves the best performance of all prior work. Our proposed FAST schemes outperform MSFP-12 by more than 2$\times$ by using lower mantissa and exponent bitwidths and switching to a higher precision in the later stage of training. 

Figure~\ref{fig:asic-results} depicts the normalized training time and energy cost for all evaluation DNNs to reach a target accuracy or BLEU. We note that performance trend and performance gain of FAST-Adaptive are consistent across models, with prior reduced/mixed precision formats outperforming FP32 by a factor of 2-3$\times$ and our proposed BFP formats achieving an additional 2-3$\times$ improvement.

\section{Conclusion}
The FAST system proposed in this paper uses block floating point (BFP) to support low-precision arithmetic to reduce DNN training time, power consumption, and hardware requirements. With FAST, we exploit an observation that earlier layers and training iterations can afford larger error margins, making them amenable to efficient low-precision computation. 

We empirically demonstrate a 2-6$\times$ speedup in training over prior work based on mixed-precision or BFP number systems while achieving similar accuracy. FAST's superior performance is due to our architectural choice of using the BFP number system, use of stochastic rounding in BFP, and modular fMAC design to support multiple precisions. This work shows that variable precison BFP with stochastic rounding offers a promising strategy in speeding up training and improving its efficiency. As DNN training is now often distributed across multi-chip systems~\cite{jouppi2020domain,tesladojo}, future work is to study how well FAST could scale in such a multi-chip deployment.



\nocite{casey2009single}
\newpage
\newpage

\bibliographystyle{IEEEtranS}
\bibliography{refs}

\begin{thebibliography}{10}
\providecommand{\url}[1]{#1}
\csname url@samestyle\endcsname
\providecommand{\newblock}{\relax}
\providecommand{\bibinfo}[2]{#2}
\providecommand{\BIBentrySTDinterwordspacing}{\spaceskip=0pt\relax}
\providecommand{\BIBentryALTinterwordstretchfactor}{4}
\providecommand{\BIBentryALTinterwordspacing}{\spaceskip=\fontdimen2\font plus
\BIBentryALTinterwordstretchfactor\fontdimen3\font minus
  \fontdimen4\font\relax}
\providecommand{\BIBforeignlanguage}[2]{{%
\expandafter\ifx\csname l@#1\endcsname\relax
\typeout{** WARNING: IEEEtranS.bst: No hyphenation pattern has been}%
\typeout{** loaded for the language `#1'. Using the pattern for}%
\typeout{** the default language instead.}%
\else
\language=\csname l@#1\endcsname
\fi
#2}}
\providecommand{\BIBdecl}{\relax}
\BIBdecl

\bibitem{iwslt-benchmark}
``Iwslt2014 benchmark,''
  \url{https://paperswithcode.com/sota/machine-translation-on-iwslt2014-german}.

\bibitem{pytorch-settings}
``Pytorch official implementation for imagenet,''
  \url{https://github.com/pytorch/examples/blob/master/imagenet/main.py}.

\bibitem{tesladojo}
``Tesla dojo technology,''
  \url{https://tesla-cdn.thron.com/static/SBY4B9_tesla-dojo-technology_OPNZ0M.pdf?xseo=&response-content-disposition=inline%3Bfilename%3D%22tesla-dojo-technology.pdf%22},
  2021.

\bibitem{cacti}
``Cacti: An integrated cache and memory access time, cycle time, area, leakage,
  and dynamic power model,'' \url{https://github.com/HewlettPackard/cacti}.

\bibitem{synopsysdesigncompiler}
``Design compiler: Rtl synthesis,''
  \url{https://www.synopsys.com/support/training/rtl-synthesis/design-compiler-rtl-synthesis.html}.

\bibitem{nangatelib}
``Nangate freepdk45 open cell library,''
  \url{http://www.nangate.com/?page_id=2325}.

\bibitem{banner2018scalable}
R.~Banner, I.~Hubara, E.~Hoffer, and D.~Soudry, ``Scalable methods for 8-bit
  training of neural networks,'' in \emph{NeurIPS}, 2018, pp. 5151--5159.

\bibitem{bilaniuk2019bit}
O.~Bilaniuk, S.~Wagner, Y.~Savaria, and J.-P. David, ``Bit-slicing fpga
  accelerator for quantized neural networks,'' in \emph{2019 IEEE International
  Symposium on Circuits and Systems (ISCAS)}.\hskip 1em plus 0.5em minus
  0.4em\relax IEEE, 2019, pp. 1--5.

\bibitem{casey2009single}
M.~C. Casey, ``Single-event effects in digital cmos circuits operating at
  ultra-low power,'' Ph.D. dissertation, Vanderbilt University, 2009,
  \url{https://www.researchgate.net/publication/224567215_Single-Event_Effects_on_Ultra-Low_Power_CMOS_Circuits}.

\bibitem{chmiel2020neural}
B.~Chmiel, L.~Ben-Uri, M.~Shkolnik, E.~Hoffer, R.~Banner, and D.~Soudry,
  ``Neural gradients are near-lognormal: improved quantized and sparse
  training,'' 2020.

\bibitem{choi2020energy}
S.~Choi, J.~Sim, M.~Kang, Y.~Choi, H.~Kim, and L.-S. Kim, ``An energy-efficient
  deep convolutional neural network training accelerator for in situ
  personalization on smart devices,'' \emph{IEEE Journal of Solid-State
  Circuits}, vol.~55, no.~10, pp. 2691--2702, 2020.

\bibitem{coleman2017dawnbench}
C.~Coleman, D.~Narayanan, D.~Kang, T.~Zhao, J.~Zhang, L.~Nardi, P.~Bailis,
  K.~Olukotun, C.~R{\'e}, and M.~Zaharia, ``Dawnbench: An end-to-end deep
  learning benchmark and competition,'' \emph{Training}, vol. 100, no. 101, p.
  102, 2017.

\bibitem{courbariaux2014training}
M.~Courbariaux, Y.~Bengio, and J.-P. David, ``Training deep neural networks
  with low precision multiplications,'' \emph{arXiv preprint arXiv:1412.7024},
  2014.

\bibitem{courbariaux2015binaryconnect}
M.~Courbariaux, Y.~Bengio, and J.-P. David, ``Binaryconnect: Training deep
  neural networks with binary weights during propagations,'' in \emph{Advances
  in neural information processing systems}, 2015, pp. 3123--3131.

\bibitem{rouhani20bfp}
\BIBentryALTinterwordspacing
B.~Darvish~Rouhani, D.~Lo, R.~Zhao, M.~Liu, J.~Fowers, K.~Ovtcharov,
  A.~Vinogradsky, S.~Massengill, L.~Yang, R.~Bittner, A.~Forin, H.~Zhu, T.~Na,
  P.~Patel, S.~Che, L.~Chand~Koppaka, X.~SONG, S.~Som, K.~Das, S.~T,
  S.~Reinhardt, S.~Lanka, E.~Chung, and D.~Burger, ``Pushing the limits of
  narrow precision inferencing at cloud scale with microsoft floating point,''
  in \emph{Advances in Neural Information Processing Systems}, H.~Larochelle,
  M.~Ranzato, R.~Hadsell, M.~F. Balcan, and H.~Lin, Eds., vol.~33.\hskip 1em
  plus 0.5em minus 0.4em\relax Curran Associates, Inc., 2020, pp.
  10\,271--10\,281. [Online]. Available:
  \url{https://proceedings.neurips.cc/paper/2020/file/747e32ab0fea7fbd2ad9ec03daa3f840-Paper.pdf}
\BIBentrySTDinterwordspacing

\bibitem{deng2009imagenet}
J.~Deng, W.~Dong, R.~Socher, L.-J. Li, K.~Li, and L.~Fei-Fei, ``Imagenet: A
  large-scale hierarchical image database,'' in \emph{Computer Vision and
  Pattern Recognition, 2009. CVPR 2009. IEEE Conference on}.\hskip 1em plus
  0.5em minus 0.4em\relax IEEE, 2009, pp. 248--255.

\bibitem{drumondbfp18}
M.~Drumond, T.~Lin, M.~Jaggi, and B.~Falsafi, ``Training dnns with hybrid block
  floating point,'' in \emph{Proceedings of the 32nd International Conference
  on Neural Information Processing Systems}, ser. NIPS'18.\hskip 1em plus 0.5em
  minus 0.4em\relax Red Hook, NY, USA: Curran Associates Inc., 2018, p.
  451–461.

\bibitem{everingham2011pascal}
M.~Everingham and J.~Winn, ``The pascal visual object classes challenge 2012
  (voc2012) development kit,'' \emph{Pattern Analysis, Statistical Modelling
  and Computational Learning, Tech. Rep}, vol.~8, p.~5, 2011.

\bibitem{fleischer18}
B.~Fleischer, S.~Shukla, M.~Ziegler, J.~Silberman, J.~Oh, V.~Srinivasan,
  J.~Choi, S.~Mueller, A.~Agrawal, T.~Babinsky, N.~Cao, C.-Y. Chen, P.~Chuang,
  T.~Fox, G.~Gristede, M.~Guillorn, H.~Haynie, M.~Klaiber, D.~Lee, S.-H. Lo,
  G.~Maier, M.~Scheuermann, S.~Venkataramani, C.~Vezyrtzis, N.~Wang, F.~Yee,
  C.~Zhou, P.-F. Lu, B.~Curran, L.~Chang, and K.~Gopalakrishnan, ``A scalable
  multi- teraops deep learning processor core for ai trainina and inference,''
  in \emph{2018 IEEE Symposium on VLSI Circuits}, 2018, pp. 35--36.

\bibitem{fowers2018brainwave}
\BIBentryALTinterwordspacing
J.~Fowers, K.~Ovtcharov, M.~Papamichael, T.~Massengill, M.~Liu, D.~Lo,
  S.~Alkalay, M.~Haselman, L.~Adams, M.~Ghandi, S.~Heil, P.~Patel, A.~Sapek,
  G.~Weisz, L.~Woods, S.~Lanka, S.~K. Reinhardt, A.~M. Caulfield, E.~S. Chung,
  and D.~Burger, ``A configurable cloud-scale dnn processor for real-time ai,''
  in \emph{Proceedings of the 45th Annual International Symposium on Computer
  Architecture}, ser. ISCA '18.\hskip 1em plus 0.5em minus 0.4em\relax IEEE
  Press, 2018, p. 1–14. [Online]. Available:
  \url{https://doi.org/10.1109/ISCA.2018.00012}
\BIBentrySTDinterwordspacing

\bibitem{bfloat}
``Bfloat16: The secret to high performance on cloud tpus,''
  \url{https://cloud.google.com/blog/products/ai-machine-learning/bfloat16-the-secret-to-high-performance-on-cloud-tpus},
  Google, accessed: 2021-03-29.

\bibitem{gupta2015deep}
S.~Gupta, A.~Agrawal, K.~Gopalakrishnan, and P.~Narayanan, ``Deep learning with
  limited numerical precision,'' in \emph{International Conference on Machine
  Learning}, 2015, pp. 1737--1746.

\bibitem{hubara2017quantized}
I.~Hubara, M.~Courbariaux, D.~Soudry, R.~El-Yaniv, and Y.~Bengio, ``Quantized
  neural networks: Training neural networks with low precision weights and
  activations,'' \emph{The Journal of Machine Learning Research}, vol.~18,
  no.~1, pp. 6869--6898, 2017.

\bibitem{jacob2018quantization}
B.~Jacob, S.~Kligys, B.~Chen, M.~Zhu, M.~Tang, A.~Howard, H.~Adam, and
  D.~Kalenichenko, ``Quantization and training of neural networks for efficient
  integer-arithmetic-only inference,'' in \emph{Proceedings of the IEEE
  Conference on Computer Vision and Pattern Recognition}, 2018, pp. 2704--2713.

\bibitem{jouppi2020domain}
N.~P. Jouppi, D.~H. Yoon, G.~Kurian, S.~Li, N.~Patil, J.~Laudon, C.~Young, and
  D.~Patterson, ``A domain-specific supercomputer for training deep neural
  networks,'' \emph{Communications of the ACM}, vol.~63, no.~7, pp. 67--78,
  2020.

\bibitem{judd2016stripes}
P.~Judd, J.~Albericio, T.~Hetherington, T.~M. Aamodt, and A.~Moshovos,
  ``Stripes: Bit-serial deep neural network computing,'' in
  \emph{Microarchitecture (MICRO), 2016 49th Annual IEEE/ACM International
  Symposium on}.\hskip 1em plus 0.5em minus 0.4em\relax IEEE, 2016, pp. 1--12.

\bibitem{kapur2017low}
S.~Kapur, A.~Mishra, and D.~Marr, ``Low precision rnns: Quantizing rnns without
  losing accuracy,'' \emph{arXiv preprint arXiv:1710.07706}, 2017.

\bibitem{kingma2014adam}
D.~P. Kingma and J.~Ba, ``Adam: A method for stochastic optimization,''
  \emph{arXiv preprint arXiv:1412.6980}, 2014.

\bibitem{koster2017flexpoint}
U.~K{\"o}ster, T.~J. Webb, X.~Wang, M.~Nassar, A.~K. Bansal, W.~H. Constable,
  O.~H. Elibol, S.~Gray, S.~Hall, L.~Hornof \emph{et~al.}, ``Flexpoint: An
  adaptive numerical format for efficient training of deep neural networks,''
  \emph{arXiv preprint arXiv:1711.02213}, 2017.

\bibitem{kung1982systolic}
H.~T. Kung, ``Why systolic architectures?'' \emph{IEEE Computer}, vol.~15, pp.
  37--46, 1982.

\bibitem{kung2019packing}
H.~T. Kung, B.~McDanel, and S.~Q. Zhang, ``Packing sparse convolutional neural
  networks for efficient systolic array implementations: Column combining under
  joint optimization,'' \emph{24th ACM International Conference on
  Architectural Support for Programming Languages and Operating Systems}, 2019.

\bibitem{lee20197}
J.~Lee, J.~Lee, D.~Han, J.~Lee, G.~Park, and H.-J. Yoo, ``7.7 lnpu: A 25.3
  tflops/w sparse deep-neural-network learning processor with fine-grained
  mixed precision of fp8-fp16,'' in \emph{2019 IEEE International Solid-State
  Circuits Conference-(ISSCC)}.\hskip 1em plus 0.5em minus 0.4em\relax IEEE,
  2019, pp. 142--144.

\bibitem{mahmoud2020tensordash}
M.~Mahmoud, I.~Edo, A.~H. Zadeh, O.~M. Awad, G.~Pekhimenko, J.~Albericio, and
  A.~Moshovos, ``Tensordash: Exploiting sparsity to accelerate deep neural
  network training,'' in \emph{2020 53rd Annual IEEE/ACM International
  Symposium on Microarchitecture (MICRO)}.\hskip 1em plus 0.5em minus
  0.4em\relax IEEE, 2020, pp. 781--795.

\bibitem{micikevicius2018mixed}
\BIBentryALTinterwordspacing
P.~Micikevicius, S.~Narang, J.~Alben, G.~Diamos, E.~Elsen, D.~Garcia,
  B.~Ginsburg, M.~Houston, O.~Kuchaiev, G.~Venkatesh, and H.~Wu, ``Mixed
  precision training,'' in \emph{International Conference on Learning
  Representations}, 2018. [Online]. Available:
  \url{https://openreview.net/forum?id=r1gs9JgRZ}
\BIBentrySTDinterwordspacing

\bibitem{neelakantan2015adding}
A.~Neelakantan, L.~Vilnis, Q.~V. Le, I.~Sutskever, L.~Kaiser, K.~Kurach, and
  J.~Martens, ``Adding gradient noise improves learning for very deep
  networks,'' \emph{arXiv preprint arXiv:1511.06807}, 2015.

\bibitem{tf32}
``Accelerating ai training with nvidia tf32 tensor cores,''
  \url{https://developer.nvidia.com/blog/accelerating-ai-training-with-tf32-tensor-cores/},
  Nvidia, accessed: 2021-03-29.

\bibitem{lee2020}
J.~Oh, S.~Lee, M.~Kang, M.~Ziegler, J.~Silberman, A.~Agrawal, S.~Venkataramani,
  B.~Fleischer, M.~Guillorn, J.~Choi, W.~Wang, S.~Mueller, S.~Ben-Yehuda,
  J.~Bonanno, N.~Cao, R.~Casatuta, C.~Chen, M.~Cohen, O.~Erez, T.~Fox,
  G.~Gristede, H.~Haynie, V.~Ivanov, S.~Koswatta, S.~Lo, M.~Lutz, G.~Maier,
  A.~Mesh, Y.~Nustov, S.~Rider, M.~Schaal, M.~Scheuermann, X.~Sun, N.~Wang,
  F.~Yee, C.~Zhou, V.~Shah, B.~Curran, V.~Srinivasan, P.~Lu, S.~Shukla,
  K.~Gopalakrishnan, and L.~Chang, ``\BIBforeignlanguage{English}{A 3.0 tflops
  0.62v scalable processor core for high compute utilization ai training and
  inference},'' in \emph{\BIBforeignlanguage{English}{2020 IEEE Symposium on
  VLSI Circuits, VLSI Circuits 2020 - Proceedings}}, ser. IEEE Symposium on
  VLSI Circuits, Digest of Technical Papers.\hskip 1em plus 0.5em minus
  0.4em\relax Institute of Electrical and Electronics Engineers Inc., Jun.
  2020, publisher Copyright: {\textcopyright} 2020 IEEE.; null ; Conference
  date: 16-06-2020 Through 19-06-2020.

\bibitem{park2017weighted}
E.~Park, J.~Ahn, and S.~Yoo, ``Weighted-entropy-based quantization for deep
  neural networks,'' in \emph{Proceedings of the IEEE Conference on Computer
  Vision and Pattern Recognition}, 2017, pp. 5456--5464.

\bibitem{pillmeier2002design}
M.~R. Pillmeier, M.~J. Schulte, and E.~G. Walters~III, ``Design alternatives
  for barrel shifters,'' in \emph{Advanced Signal Processing Algorithms,
  Architectures, and Implementations XII}, vol. 4791.\hskip 1em plus 0.5em
  minus 0.4em\relax International Society for Optics and Photonics, 2002, pp.
  436--447.

\bibitem{qin2020}
E.~Qin, A.~Samajdar, H.~Kwon, V.~Nadella, S.~Srinivasan, D.~Das, B.~Kaul, and
  T.~Krishna, ``Sigma: A sparse and irregular gemm accelerator with flexible
  interconnects for dnn training,'' in \emph{2020 IEEE International Symposium
  on High Performance Computer Architecture (HPCA)}, 2020, pp. 58--70.

\bibitem{redmon2017yolo9000}
J.~Redmon and A.~Farhadi, ``Yolo9000: better, faster, stronger,'' in
  \emph{Proceedings of the IEEE conference on computer vision and pattern
  recognition}, 2017, pp. 7263--7271.

\bibitem{sun2019hybrid}
X.~Sun, J.~Choi, C.-Y. Chen, N.~Wang, S.~Venkataramani, V.~V. Srinivasan,
  X.~Cui, W.~Zhang, and K.~Gopalakrishnan, ``Hybrid 8-bit floating point (hfp8)
  training and inference for deep neural networks,'' \emph{Advances in neural
  information processing systems}, vol.~32, pp. 4900--4909, 2019.

\bibitem{Wilkinson_64}
J.~H. Wilkinson, \emph{Rounding Errors in Algebraic Processes}.\hskip 1em plus
  0.5em minus 0.4em\relax Dover Publications, 1964.

\bibitem{yang2020procrustes}
D.~Yang, A.~Ghasemazar, X.~Ren, M.~Golub, G.~Lemieux, and M.~Lis, ``Procrustes:
  a dataflow and accelerator for sparse deep neural network training,'' in
  \emph{2020 53rd Annual IEEE/ACM International Symposium on Microarchitecture
  (MICRO)}.\hskip 1em plus 0.5em minus 0.4em\relax IEEE, 2020, pp. 711--724.

\bibitem{zhang2019eager}
J.~Zhang, X.~Chen, M.~Song, and T.~Li, ``Eager pruning: algorithm and
  architecture support for fast training of deep neural networks,'' in
  \emph{2019 ACM/IEEE 46th Annual International Symposium on Computer
  Architecture (ISCA)}.\hskip 1em plus 0.5em minus 0.4em\relax IEEE, 2019, pp.
  292--303.

\bibitem{zhu2016trained}
C.~Zhu, S.~Han, H.~Mao, and W.~J. Dally, ``Trained ternary quantization,''
  \emph{arXiv preprint arXiv:1612.01064}, 2016.

\end{thebibliography}

\end{document}